%% file: arxiv_main.tex
\documentclass{article} 
\usepackage{iclr2027_conference,times}

\input{math_commands.tex}

\usepackage[utf8]{inputenc} 
\usepackage[T1]{fontenc}    
\usepackage{hyperref}       
\usepackage{url}            
\usepackage{booktabs}       
\usepackage{amsfonts}       
\usepackage{nicefrac}       
\usepackage{microtype}      
\usepackage{xcolor}         
\usepackage{natbib}
\usepackage{tikz}
\usepackage{verbatim}
\usepackage{fancyvrb}
\usepackage{subcaption}
\usepackage{amssymb}
\usepackage{amsmath}
\usepackage{bbm}
\usepackage{enumitem}
\usetikzlibrary{arrows.meta,calc,positioning}

\PassOptionsToPackage{disable}{todonotes}
\input{macros}

\title{The Belief--Action Gap in Language Model Agents}

\author{%
\parbox[t]{\dimexpr\textwidth-2\tabcolsep\relax}{%
\raggedright
\normalfont\normalsize
\textbf{Ayse Sila Okcu}\textsuperscript{1}\thanks{These authors contributed equally and are listed alphabetically by surname.}\kern0.4em,
\textbf{Weisheng Wang}\textsuperscript{2\ensuremath{\ast}},
\textbf{Tongmeng Xie}\textsuperscript{3\ensuremath{\ast}},
\textbf{Niall Dalton}\textsuperscript{4}\\[0.4ex]
\textbf{Angelos Nalmpantis}\textsuperscript{5},
\textbf{Raghu Arghal}\textsuperscript{6},
\textbf{Mario Giulianelli}\textsuperscript{4}\\[1ex]
\small
\textsuperscript{1} University of Cambridge\\
\textsuperscript{2} Utrecht University\\
\textsuperscript{3} University of York\\
\textsuperscript{4} University College London, University of London\\
\textsuperscript{5} TKH AI Technology\\
\textsuperscript{6} University of Pennsylvania
}}

\hypersetup{%
  hidelinks,
  pdftitle={The Belief--Action Gap in Language Model Agents},
  pdfauthor={Ayse Sila Okcu, Weisheng Wang, Tongmeng Xie, Niall Dalton, Angelos Nalmpantis, Raghu Arghal, Mario Giulianelli}
}

\iclrfinalcopy
\begin{document}

\maketitle

\fancyhead{}
\lhead{Preprint}
\renewcommand{\headrulewidth}{0pt}

\begin{abstract}
Language model (LM) agents can fail for different reasons, requiring different explanations and interventions. We study how state beliefs, action values, and action preferences interact as LM agents reason and commit to decisions in text-rendered grid worlds. We estimate task-relevant beliefs using report-based and representation-based probes, and action values using inverse reinforcement learning. We find that task-relevant state and value information is often represented more reliably than it is reflected in behaviour. Tracking action preferences throughout reasoning further shows that decisions sharpen around a commitment point and reported belief uncertainty decreases near commitment, although individual belief changes do not consistently explain shifts in action preference. 
Finally, forced actions follow reasoning prefixes from other decisions when transplanted, whereas the tested reasoning-window activation edits provide limited control over the action. 
Together, these results suggest that some failures arise not from missing task-relevant information, but from how available information is translated into action. Our measurements distinguish information availability from belief-action consistency without by themselves identifying the cause of each failure.
Our code is available at: \url{https://anonymous.4open.science/r/belief-action-gap-F745/}
\end{abstract}



\section{Introduction}
\label{sec:introduction}
As language models are increasingly used as agents that observe, reason about, and act within external environments, evaluation based on outputs alone cannot diagnose why an agent fails \citep{wang2024survey}.
A suboptimal action may arise because the agent represents the current state incorrectly, because its preferences favour an undesirable successor state, or because available state and preference information fails to determine the selected action. These failure modes can produce similar behaviour while requiring different explanations and interventions \citep{amodei2016concrete,schmied2026llms}.

We therefore analyse agent decisions in terms of three components: the agent's \emph{belief} about the current environment state, its \emph{preference structure} over possible outcomes, and a \emph{decision mechanism} that translates these beliefs and preferences into a selected action. 
This decomposition allows us to distinguish failures of state estimation from failures of preference, as well as cases in which relevant information is available but does not govern behaviour.

Existing work provides tools for studying parts of this process. Probing methods show that language model activations can encode spatial and other task-relevant information about the state of their environment \citep{li2022emergent,nanda2023emergent,gurnee2024language,arghal2026behavioural}. Successful decoding, however, establishes information availability rather than functional use in behaviour \citep{hewitt2019designing,belinkov2022probing,orgad2025llms}. Behavioural methods such as inverse reinforcement learning \citep{ng2000algorithms,ziebart2008maximum} can infer latent preferences from observed choices, while related work studies preferences in language-model agents through feedback and behavioural biases \citep{gao2024aligning,lu2025aligning,khan2026agents}. Decision-theoretic work on the value of information formalises the cost of uncertainty for action selection \citep{howard1966information}. Chain-of-thought traces provide a useful view of decision formation but need not faithfully reflect the computations responsible for the final answer \citep{turpin2023language,lanham2023measuring}. Recent work also tracks how answer preferences, confidence, and uncertainty evolve during reasoning \citep{catala2026answer,hosseini2026early,caldarella2026thinking}, while other work decodes future responses from model internals \citep{belrose2023eliciting,pal2023future,dong2025emergent} and tests causal influence through activation interventions \citep{turner2023steering,zhang2024towards}. Decodability or localisation also need not identify representations that provide behavioural control \citep{hase2023does}. Together, these methods do not by themselves explain where representation fails to translate into action.


We study how beliefs and preferences are transformed into action in a language model agent operating in a text-rendered grid world with known states, transitions, and optimal actions. We estimate task-relevant beliefs from internal activations and explicit reports, infer state preferences from behaviour using inverse reinforcement learning, track action preferences throughout reasoning, and intervene on both reasoning text and hidden representations. Together, these analyses test when task-relevant information is available, how action preferences form, and whether that information causally influences the selected action.

Our main contributions are:
\begin{itemize}[
leftmargin=1.4em,
label=\raisebox{0.35ex}{\rule{0.75ex}{0.75ex}},
labelsep=0.5em
]
\item \textbf{Belief--action diagnostics.}
We combine estimated state beliefs and inferred preferences to measure whether emitted actions are consistent with the information available to the agent.

\item \textbf{Decision formation during reasoning.}
We track evolving action preferences and identify a commitment point at which the preferred action stabilises, then examine how beliefs and hidden-state patterns change around this transition.

\item \textbf{Causal tests of information use.}
We show that reasoning-text interventions can strongly redirect action preferences, while targeted edits to highly decodable hidden representations are much less effective, separating information availability from causal control.

\item \textbf{Empirical evidence for a belief--action gap.}
We find that task-relevant information can remain available even when behaviour is suboptimal, and that improved action consistency after reasoning does not simply result from more precise state estimation.
\end{itemize}


\section{Estimating Beliefs and Values}
\label{sec:estimating}


An agent may fail by misidentifying its state, preferring the wrong outcome, or acting inconsistently with its available information. We separate these possibilities using an estimated state belief $\hat q_t(s)$, an inferred state-value function $g_\theta(s)$, and the emitted action $a_t$. Beliefs come from activation probes; values are inferred from observed choices using one-step inverse reinforcement learning. Together, they yield expected local regret, which we decompose into the belief-–action gap and the expected value of perfect state information. We measure decoded belief uncertainty with Shannon entropy.

\subsection{Grid World Environment}
\label{sec:env-data}

\begin{figure}[htbp]
\centering

\begin{subfigure}[c]{0.3\textwidth}
\centering
\begin{BVerbatim}
  0 1 2 3 4 5 6 7 8
0 # # # # # # # # #
1 # . . . # . . . #
2 # . . . D . . . #
3 # . . . # . G . #
4 # . # # # # # # #
5 # . . . . . . . #
6 # . . . # . K . #
7 # . . A # . . . #
8 # # # # # # # # #
\end{BVerbatim}

\caption{}
\label{fig:grid_text}
\end{subfigure}
\hspace{0.02\textwidth}
\begin{subfigure}[c]{0.27\textwidth}
\centering
\includegraphics[
    width=\linewidth,
    trim=220 100 220 100,
    clip
]{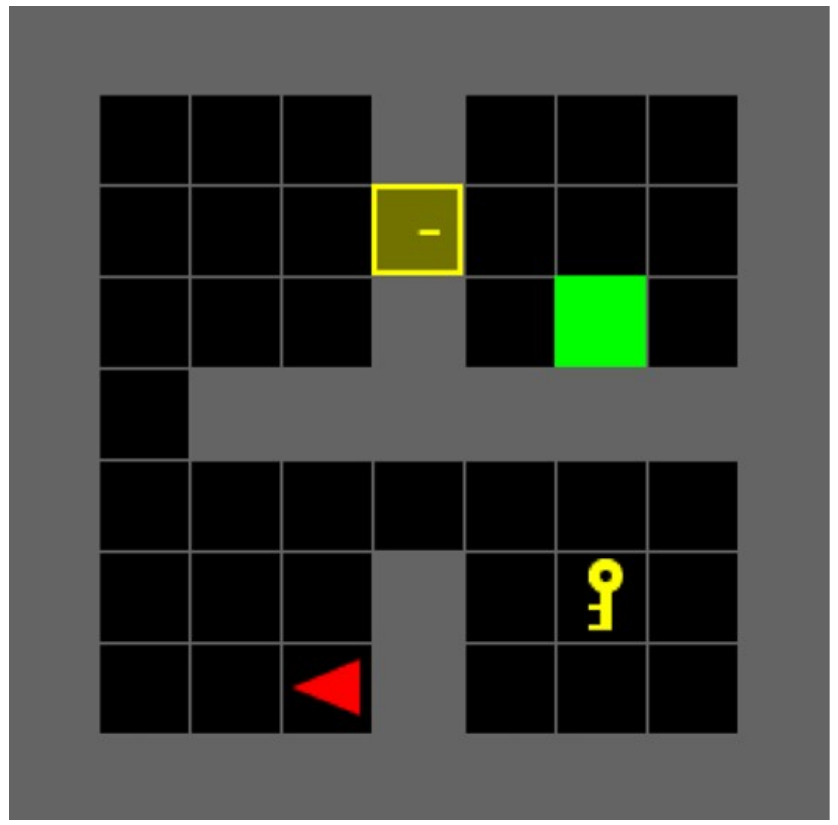}
\caption{}
\label{fig:grid_visual}
\end{subfigure}
\caption{
Example DoorKey environment shown in text and spatial form.
The language model receives the text-rendered grid in \textbf{(a)}, while \textbf{(b)} visualises the same underlying environment used to define states, transitions, and planner-optimal actions.
$A$ denotes the agent, $K$ the key, $D$ the door, $G$ the goal, and filled cells denote walls.
}
\label{fig:environment}
\end{figure}

We study a language model agent in a fully observable, text-rendered grid world, following the setup of \citet{arghal2026behavioural}. 
The grid contains walls, open cells, a goal cell, a locked door that blocks the route to the goal, and a key that the agent must collect before passing through the door. 
The key is collected automatically when the agent enters its cell, after which the door can be unlocked. Transitions are deterministic, with a valid movement changing the agent's position by one cell, while an action directed towards a wall or locked door leaves the agent in its current position. Figure~\ref{fig:environment} shows an example grid.
For a grid layout $G$, we represent the environment state at decision step $t$ as $s_t=(x_t,y_t,h_t,d_t)$.
$(x_t,y_t)$ are the agent's grid coordinates, $h_t \in \{0,1\}$ indicates whether it is carrying the key, and $d_t \in \{0,1\}$ indicates whether the door is open. 
The layout $G$ specifies the fixed locations of walls, the key, the door, and the goal.
For our analyses, we use a fixed $9\times9$ grid layout. 
The grid contains 39 walkable cells and admits 142 valid task states.
At each step, the language model agent receives the fully-observable grid, including coordinate labels, and its key-possession status, but no previous environment states or actions, making the interaction Markovian with respect to the task state. It then produces a sequence of reasoning tokens and emits one action from $\mathcal A=\{\textsc{left},\textsc{right},\textsc{up},\textsc{down}\}$.

\subsection{Estimating Beliefs}
\label{sec:estimating-beliefs}

We estimate the agent's task-state beliefs using both representation-based and report-based estimates. Representation-based beliefs are obtained from probes trained on internal activations \citep{alain2016understanding,gurnee2024language}. For a token region $p$ and model layer $\ell$, the probe input $\mathbf{u}_{t}^{(\ell,p)}$ consists of the activations from that region at decision step $t$. We train separate probes for the three components of the task state $s_t$; the agent position $(x,y)$, key possession $h$, and door status $d$. Position is treated as a single categorical variable over grid locations, while key possession and door status are binary. Their predicted distributions are combined into a factorised distribution over states and renormalised over the valid state set $\mathcal S_G$, giving the representation-based belief $\widehat q_{t,p}(s)$.

We also obtain report-based estimates by directly querying the model about task-relevant features of the same environment state. We evaluate report accuracy using the highest-probability answer and measure uncertainty using Shannon entropy. Because these reports are elicited in separate forward passes, they are analysed separately and are not used to compute the belief--action gap.
Full probe definitions and training details are provided in
Appendix~\ref{app:representation-beliefs}, and the report-based protocol is described in Appendix~\ref{app:report-beliefs}.

\subsection{Estimating Values}
\label{sec:estimating-values}

We infer the agent's state preferences from observed state--action pairs using a one-step maximum-entropy inverse reinforcement learning model \citep{ng2000algorithms,ziebart2008maximum}. Each state $s$ is represented by an activation-derived feature vector $\boldsymbol{\phi}_p(s)$ obtained by averaging the corresponding model representations across occurrences of that state. We fit separate models for the pre-reasoning and post-reasoning regions. The state-value function is linear in this representation:
$g_{\theta_p} (s)
=
\theta^\top \boldsymbol{\phi}_p(s).$
Since transitions are deterministic, an action is evaluated through the value of its successor state $T_G(s,a)$. We convert these successor values into action probabilities using a Boltzmann policy,
\begin{align}
    p_\theta(a\mid s,G)
    &=
    \frac{
        \exp\!\left(
            \beta g_\theta(T_G(s,a))
        \right)
    }{
        \sum_{a'\in\mathcal A}
        \exp\!\left(
            \beta g_\theta(T_G(s,a'))
        \right)
    },
    \label{eq:irl-policy}
\end{align}
We then fit $\theta$ to the actions emitted by the language model by minimising their negative log-likelihood.
We call this model one-step IRL as actions are compared through their immediate successor states rather than explicit trajectory rollouts. We interpret $g_\theta(\boldsymbol{\phi}_p(s))$ as the agent's inferred subjective value of state $s$. Further details on the state representations and model fitting are provided in
Appendix~\ref{app:irl-details}.

\subsection{Quantifying Belief--Action Consistency}
\label{sec:belief-action-gap}

We combine the estimated beliefs and values to characterise how the emitted action relates to agent's inferred state belief and value function. For a known state $s$, we define the regret of action $a$ as
\begin{align}
    \rho_{\theta}(s,a)
    &=
    \max_{a' \in \mathcal A}
    g_{\theta}\!\left(T_G(s,a')\right)
    -
    g_{\theta}\!\left(T_G(s,a)\right).
    \label{eq:state-action-regret}
\end{align}
The regret is zero when $a$ leads to a maximally valued successor state and positive otherwise.

Because the agent may be uncertain about the current state, we first consider the expected local regret of the emitted action under its estimated belief,
\begin{align}
    \Delta_{t,p}
    =
    \mathbb{E}_{s \sim \widehat q_{t,p}}
    \left[
        \rho_{\theta_p}(s,a_t)
    \right]
    =
    \mathbb{E}_{s \sim \widehat q_{t,p}}
    \left[
        \max_{a' \in \mathcal A}
        g_{\theta_p}\!\left(T_G(s,a')\right)
    \right]
    -
    \mathbb{E}_{s \sim \widehat q_{t,p}}
    \left[
        g_{\theta_p}\!\left(T_G(s,a_t)\right)
    \right].
    \label{eq:expected-local-regret}
\end{align}

This quantity compares the emitted action with the state-optimal action at each possible state and then averages the resulting regret under the estimated belief. We decompose it by adding and subtracting the value of the best single action under that belief: $\Delta_{t,p}=R_{t,p}+I_{t,p},$
where
\begin{align}
    R_{t,p}
    &=
    \max_{a' \in \mathcal A}
    \mathbb{E}_{s \sim \widehat q_{t,p}}
    \left[
        g_{\theta_p}\!\left(T_G(s,a')\right)
    \right]
    -
    \mathbb{E}_{s \sim \widehat q_{t,p}}
    \left[
        g_{\theta_p}\!\left(T_G(s,a_t)\right)
    \right],
    \label{eq:belief-action-gap}
\end{align}
and
\begin{align}
    I_{t,p}
    &=
    \mathbb{E}_{s \sim \widehat q_{t,p}}
    \left[
        \max_{a' \in \mathcal A}
        g_{\theta_p}\!\left(T_G(s,a')\right)
    \right]
    -
    \max_{a' \in \mathcal A}
    \mathbb{E}_{s \sim \widehat q_{t,p}}
    \left[
        g_{\theta_p}\!\left(T_G(s,a')\right)
    \right].
    \label{eq:perfect-state-information}
\end{align}

We refer to $R_{t,p}$ as the belief--action gap. It measures how much expected value is lost by the emitted action relative to the best single action under the agent's estimated belief. A small value therefore indicates that the emitted action is consistent with the action preferred under that belief, whereas a large value indicates that another action has substantially higher expected value.

We interpret $I_{t,p}$ as the expected value of perfect state information. It measures the expected gain that would be available if the current state were known before choosing the action. This quantity is small when resolving the remaining state uncertainty would have little effect on achievable value and large when knowing the true state would substantially improve the action choice. We also quantify uncertainty in the decoded state belief using Shannon entropy:
$
H_{t,p}
=
-\mathbb{E}_{s\sim \widehat q_{t,p}}
\left[
\log \widehat q_{t,p}(s)
\right].
$
Unlike $I_{t,p}$, which depends jointly on the belief and inferred state values, $H_{t,p}$ depends only on the dispersion of the estimated belief.
We test whether reasoning improves belief--action consistency and evaluate the changes in state uncertainty and its decision relevance, motivating our analysis of $R$, $H$, and $I$ across reasoning.


\section{Linking Beliefs and Values to Behaviour}
\label{sec:analysis1}

We use grid world as a proof-of-concept setting in which states, transitions, and optimal actions are known exactly. With the belief and value machinery defined in the previous section, we ask whether suboptimal actions are better explained by inaccurate state beliefs, inappropriate state preferences, or a failure to translate available beliefs and preferences into action. 
We analyse 200 \texttt{gpt-oss-20b} grid-world trajectories comprising 3,194 decision steps. At each step, the model generates a reasoning trace and action using stochastic decoding at temperature 0.7 \citep{agarwal2025gpt}. For pre-reasoning, we force an action directly from the environment prompt before any reasoning tokens are generated.

\subsection{Inferred State Beliefs and Preferences}
\label{sec:belief-results}

We evaluate the state probes from before and after reasoning. Pre-reasoning representations use the three final prompt tokens, while post-reasoning representations use the three final reasoning tokens immediately before action generation. We use activations from layers 7, 15, and 23 and partition states into 80\% training and 20\% held-out sets, with no state appearing in both. Further probe-training details are provided in Appendix~\ref{app:representation-beliefs}, with expanded decoding results in Appendix~\ref{app:probe-results}.


Held-out \texttt{has\_key} and \texttt{door\_open} accuracy decreases from $100.0\%$ before reasoning to $89.5\%$ and $88.6\%$ after reasoning. Location decoding also becomes less precise, with top-1 accuracy falling from $34.8\%$ to $13.1\%$ and accuracy within Manhattan distance one from $95.5\%$ to $60.8\%$. Thus, although coarse location information remains recoverable, the decoded location representation becomes substantially less precise after reasoning.

We also fit the one-step agent-supervised IRL model from \Cref{sec:estimating-values} to recover the model's inferred state preferences. The resulting value model reaches held-out action-prediction accuracies of 29.2\% before reasoning and 45.2\% after reasoning. As a control, we also train the same model using planner-optimal actions, for which task-optimal action structure is substantially more recoverable. Full IRL results, including sensitivity to the inverse-temperature parameter $\beta$, are provided in Appendix~\ref{app:irl-results} and Appendix~\ref{app:planner-irl-results}; we use $\beta=1$ throughout the main analysis. We additionally analyse report-based beliefs in
Appendix~\ref{app:report-results}. In three illustrative wall-hit failures, the model correctly reports that its chosen direction is blocked while nevertheless selecting that action.

\subsection{Belief--Action Consistency}
\label{sec:decomposition}

We combine the representation-based beliefs and agent-supervised value models using the quantities defined in \Cref{sec:belief-action-gap}. 
We compute the expected local regret $\Delta_{t,p}$ and its decomposition into the belief--action gap $R_{t,p}$ and expected value of perfect state information $I_{t,p}$ before and after reasoning. We report results over all 3,194 decisions, the 2,465-step subset where every state probe is correct in both conditions, and the 762-step held-out probe set containing states unseen during training. As shown in \Cref{tab:gap-pre-post}, reasoning decreases the belief--action gap across all three subsets while increasing both the expected value of perfect state information and decoded-belief entropy.

\begin{table}[htbp]
\centering
\small
\begin{tabular}{lcccccc}
\toprule
& \multicolumn{2}{c}{\textbf{All Steps}} &
\multicolumn{2}{c}{\textbf{All Probes Correct}} &
\multicolumn{2}{c}{\textbf{Probe Held-out}} \\
\cmidrule(lr){2-3}
\cmidrule(lr){4-5}
\cmidrule(lr){6-7}
\textbf{Quantity}
& \textbf{Pre} & \textbf{Post}
& \textbf{Pre} & \textbf{Post}
& \textbf{Pre} & \textbf{Post} \\
\midrule
Expected local regret $\Delta$
& 0.681 & 0.582 & 0.676 & 0.554 & 0.700 & 0.670 \\
Belief--action gap $R$
& 0.467 & 0.268 & 0.514 & 0.288 & 0.316 & 0.199 \\
Value of perfect state information $I$
& 0.214 & 0.314 & 0.163 & 0.266 & 0.384 & 0.472 \\
Decoded Belief entropy $H$
& 1.494 & 2.085 & 1.106 & 1.811 & 2.788 & 3.020 \\
\bottomrule
\end{tabular}
\caption{
Mean expected local regret $\Delta$, belief--action gap $R$, expected value of perfect state information $I$, and decoded belief entropy $H$ before and after reasoning. The expected local regret decomposes exactly as $\Delta=R+I$.
}
\label{tab:gap-pre-post}
\end{table}

The decomposition shows why the change in expected local regret alone understates the change in belief--action consistency. 
On the held-out set, $\Delta$ decreases only from 0.700 to 0.670 because the decrease in $R$ is partly offset by the increase in $I$. After reasoning, the emitted action is therefore more consistent with the decoded belief, while the decoded state distribution is more diffuse and resolving its uncertainty is more valuable.
\begin{figure}[htbp]
    \centering
    \includegraphics[width=0.8\textwidth]{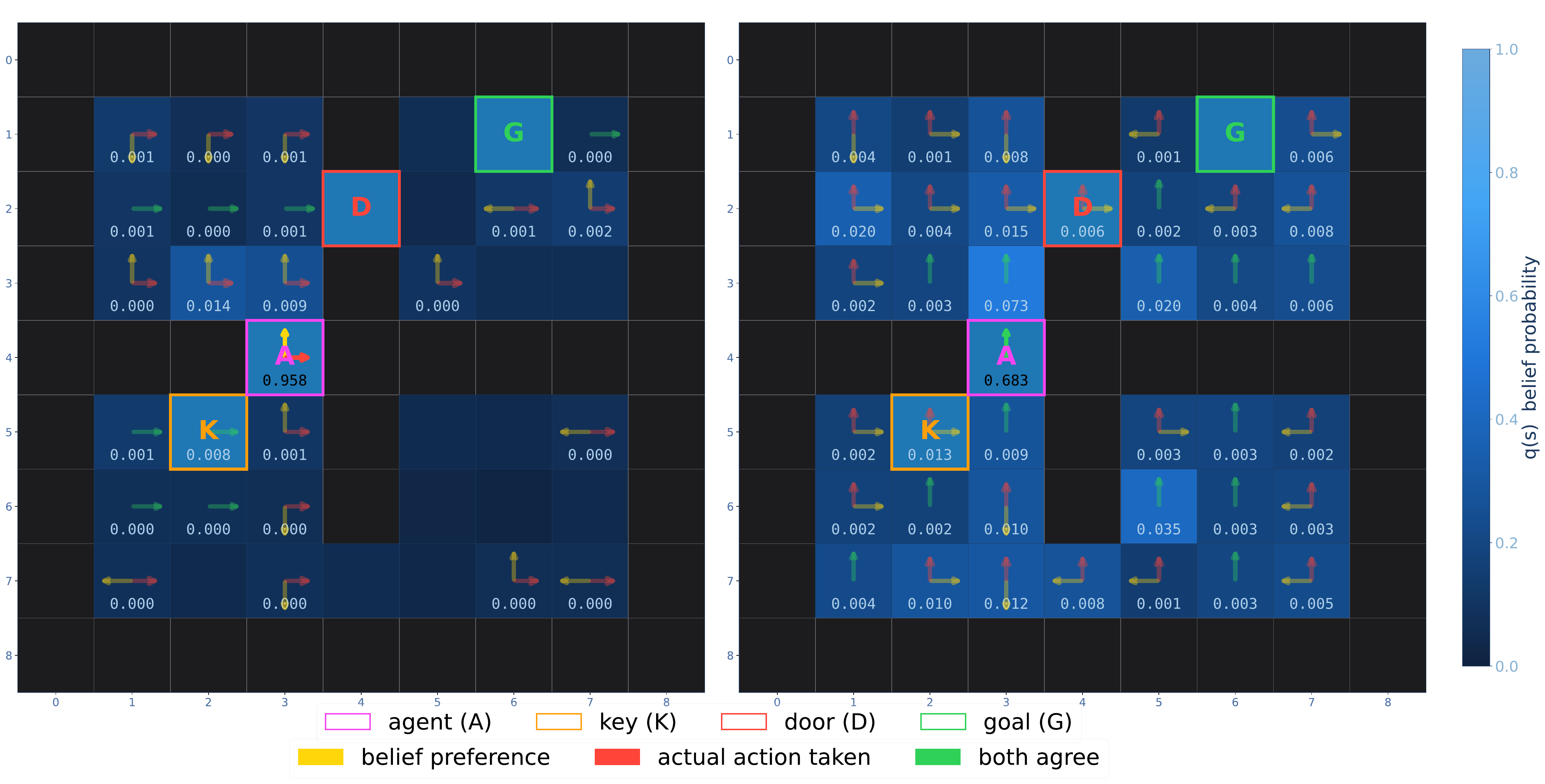}
    \caption{
    Example expected local regret under the pre-reasoning and
    post-reasoning conditions. Cell numbers
    show the probe-estimated location probabilities, with shading proportional
    to probability mass. Arrows mark the successor state preferred by the agent-supervised value model and the action actually emitted by the agent. In this
    example, reasoning increases the decoded belief entropy
    ($\operatorname{H}: 0.261 \rightarrow 1.572$) while reducing the expected local regret
    ($\mathrm{\Delta}: 2.386 \rightarrow 0.205$).
    }
    \label{fig:belief_viz_app}
\end{figure}

\Cref{fig:belief_viz_app} illustrates the same qualitative behaviour for a single decision. Before reasoning, the decoded belief is sharply concentrated while the emitted action disagrees strongly with the inferred preference model. After reasoning, the belief becomes less concentrated and the emitted action is more closely aligned with the inferred preferences.

Together, these results show that reasoning can improve belief--action consistency without producing a sharper state estimate. The decrease in $R$ indicates stronger alignment between the emitted action and the action preferred under the decoded belief, while the increases in $I$ and $H$ show that the decoded state distribution is more diffuse and that resolving this uncertainty would be more consequential for the inferred decision. The endpoint comparison, however, does not reveal when these changes arise during reasoning or whether they develop gradually or around a distinct transition.


\section{Decision Formation through Reasoning}
\label{sec:reasoning}

We examine how beliefs and action preferences evolve during reasoning using 95 trajectories comprising 1,276 decisions. At intermediate prefixes, we force a choice among the four actions and use the logits as the current action preference. Report-based beliefs are additionally available for 7,084 prefixes from 46 decisions across 31 trajectories. We use these measurements to study when action preferences stabilise and how this relates to beliefs and hidden representations.
\subsection{Formation of Action Commitment}
\label{sec:commitment}

At each evaluated prefix, we define the preferred action as the action with the
highest forced-answer probability. We define \emph{action commitment} as the
earliest prefix after which this preferred action remains unchanged for the
rest of the reasoning trace. 
Further details on prefix evaluation and commitment measurement are provided in
Appendix~\ref{app:commitment-methods}.

\begin{figure}[htbp]
\centering
\includegraphics[width=0.85\textwidth]{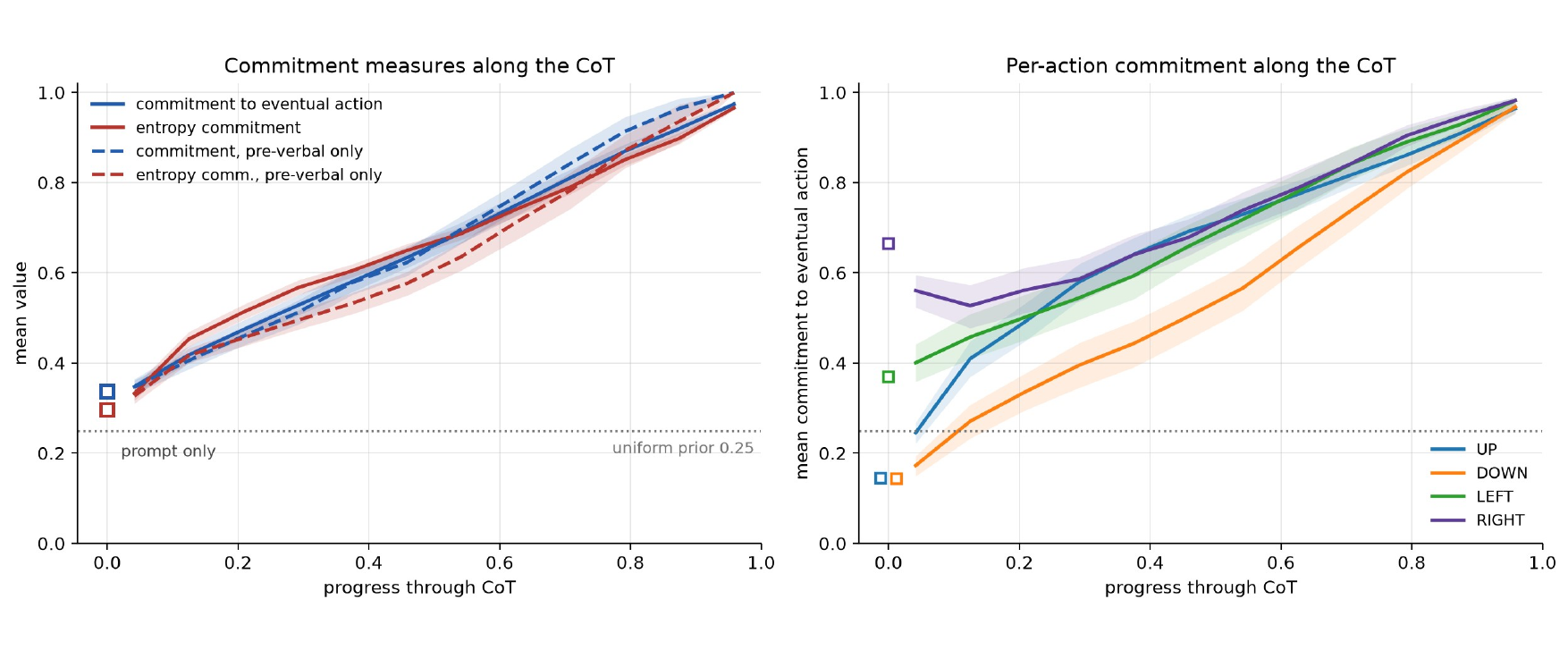}
\caption{
Evolution of action preference over reasoning
(95 trajectories, 1,276 decisions, 98,310 evaluated prefixes).
\textbf{left.} The mean probability assigned to the eventual action and
inverse normalised action entropy as reasoning progresses, together with the
same quantities before the first decisional action mention.
\textbf{right.} The probability assigned to the eventual action
separately by action. Bands are 95\% trajectory-clustered bootstrap intervals.
}
\label{fig:commitment-measures}
\end{figure}

Action preferences become more concentrated on the action
eventually emitted by the model. The mean probability assigned to the eventual
action increases from 0.35 over the first tenth of the reasoning trace to 0.92
over the final quarter. Over the same interval, inverse normalised action
entropy increases from 0.34 to 0.91. At the final evaluated prefix, the forced
action agrees with the emitted action for all decisions
(\Cref{fig:commitment-measures}).

This increase is not explained simply by the model repeating an action that it
has already mentioned. Restricting the analysis to the 35,379 prefixes before
the first decisional mention of an action produces nearly the same pattern,
with eventual-action probability increasing from 0.35 to 0.92 and inverse
normalised entropy from 0.33 to 0.89. Commitment also occurs substantially before reasoning ends. The median
commitment position is 0.49 of the reasoning trace, leaving a median 50.6\%
of the trace after the preferred action has stabilised. Among the 753 decisions
for which an explicit textual commitment can also be identified, the
forced-action preference stabilises before the textual commitment in 99.87\%
of cases, with a median lead of 523 reasoning tokens. This is consistent with
recent findings that reasoning models can settle on an answer well before
generation terminates
\citep{boppana2026reasoning,scalena2026beyond}.

Preference stabilisation is nevertheless gradual rather than a sharp transition,
consistent with recent work studying premature confidence during reasoning
\citep{gai2026understanding}.
The probability that the preferred action changes at the next evaluated prefix
falls from 0.133 in the first tenth of reasoning to 0.019 in the final tenth,
and decisions undergo 6.0 preference switches on average before commitment.
Across all evaluated prefixes, the preferred action belongs to the
planner-optimal action set in 76.7\% of cases. Changes in preference can
therefore either improve or worsen the current decision, allowing us to ask
whether such revisions are explained by corresponding changes in the model's
beliefs.

\subsection{Beliefs Around Commitment}
\label{sec:belief-commitment}

We next ask whether action commitment reflects a correction in the model's
reported understanding of the environment. We align report-based beliefs with
the commitment point for the 45 decisions that contain reports both before and
at or after commitment. Mean belief entropy decreases from 0.392 to 0.336
bits, a change of $-0.056$ with a 95\% trajectory-bootstrap interval of
$[-0.105,-0.009]$. Report accuracy, however, changes only from 82.8\% to
83.0\%, corresponding to an increase of 0.2 percentage points.

Individual changes in action preference are rarely accompanied by discrete
corrections in the reported state. Across 682 action-change events, the
correctness of at least one of the six current-state reports changes in only
14 cases, or 2.1\%. All six reports remain correct in 66 of the 161 transitions
from an optimal to a suboptimal preferred action. Thus, action stabilisation
coincides with only modest changes in reported belief accuracy.
The reports contain prospective information about future action
degradation. Among prefixes whose current preferred action is optimal, a
behavioural baseline predicts an optimality loss within the next three
sentences with AUPRC 0.273. Adding report-based belief features raises AUPRC
to 0.334, an improvement of 0.061 with a 95\% trajectory-bootstrap interval
of $[0.014,0.123]$. Thus, report-based beliefs provide predictive information
about future action degradation even though discrete belief corrections rarely
coincide with individual action changes.
Expanded transition-level results and uncertainty estimates are reported in
Appendix~\ref{app:belief-commitment}.

\subsection{Hidden States Around Decision Changes}
\label{sec:reasoning-patterns}

We finally ask whether decision revision and commitment are reflected in the
model's hidden representations. We represent each reasoning sentence using
mean-pooled activations and group these representations into
activation-defined clusters. The clustering is unsupervised, with independent
semantic annotations used only afterwards to aid interpretation. Full clustering methodology and comparisons are provided in Appendix~\ref{app:reasoning-clustering}.

The distribution of these hidden states changes around commitment. Several
clusters associated post hoc with solving and action consolidation become more
common after commitment, whereas clusters associated with reading and
reconstructing the environment become less frequent. The effect is not
explained entirely by ordinary progression through the reasoning trace. For
example, cluster \(C_{10}\) occurs in 14.0\% of true commitment sentences
compared with 5.9\% of position-matched control sentences from the same
decisions. 

Related activation patterns also appear when the preferred action itself
changes. Several clusters are significantly enriched at action-change events,
and the same clusters are enriched when the preferred action changes from
optimal to suboptimal and when it changes from suboptimal to optimal. These
states therefore appear to reflect decision revision more generally rather
than a representation specific to error detection or correction.

Taken together, the reasoning-time results show that action preference
gradually sharpens and stabilises, typically well before reasoning ends.
Reported belief uncertainty decreases modestly around commitment, but
individual belief corrections account for few changes in action preference.
Hidden-state patterns nevertheless change around decision revision and
stabilisation, suggesting that belief refinement and action commitment are
related but distinct components of the decision process.


\section{Causal Influences on Action Selection}
\label{sec:causal}

The previous sections show what information is represented and when action preferences stabilise, but not which parts of this process causally influence the selected action. We therefore intervene on the reasoning text and on a reasoning-window activation site from which action decisiveness is highly decodable. Text interventions are stratified by the current probability of the eventual action, $c_{t,k}=\pi_{t,k}(a_t)$.





\subsection{Reasoning-Text Patching}
\label{sec:text-patching}

We first test whether changing the reasoning trajectory changes the model's current action preference. The experiment uses 396 mismatched prompt--reasoning pairs, combining the prompt from a decision \(A\) with a reasoning prefix from a different decision \(B\) whose emitted action differs from that of \(A\). Prefixes are taken at approximately 10\%, 50\%, and 90\% of \(B\)'s reasoning trace, together with 396 congruent controls. Across all three cut positions, the forced-action distribution favours the action associated with the transplanted reasoning, with \(\pi(a_B)\in[0.598,0.673]\) compared with \(\pi(a_A)\in[0.129,0.165]\). Full results are given in Appendix~\ref{app:whole-chain-patching}.

A second experiment tests whether a local textual intervention is sufficient to change the current action preference. It uses 240 reasoning prefixes across six bins of current eventual-action probability \(c_{t,k}=\pi_{t,k}(a_t)\), with at most three prefixes from any one decision. At each prefix, a sentence asserting that the current runner-up action is correct is appended, together with neutral and supporting control sentences. Restricting to the 201 cases in which the injected runner-up differs from the action eventually emitted by the model, the forced top action switches to the injected action in 100\% of cuts with \(c_{t,k}<0.7\), 92.5\% for \(0.7\leq c_{t,k}<0.85\), 90.0\% for \(0.85\leq c_{t,k}<0.95\), and 62.5\% for \(c_{t,k}\geq0.95\). Susceptibility decreases with current action confidence (Spearman \(\rho=-0.421\), \(p=5.0\times10^{-10}\)) and remains significant after controlling for reasoning position (partial \(\rho=-0.175\), \(p=0.013\)).
A neutral sentence changes the preferred action in only 9.6\% of cuts, while a supporting sentence raises the mean probability of the current top action from 0.757 to 0.989. Reasoning text therefore has substantial causal influence on the current decision, although this influence weakens as the eventual action becomes more strongly preferred.

\begin{figure}[htbp]
\centering
\includegraphics[width=0.7\textwidth]{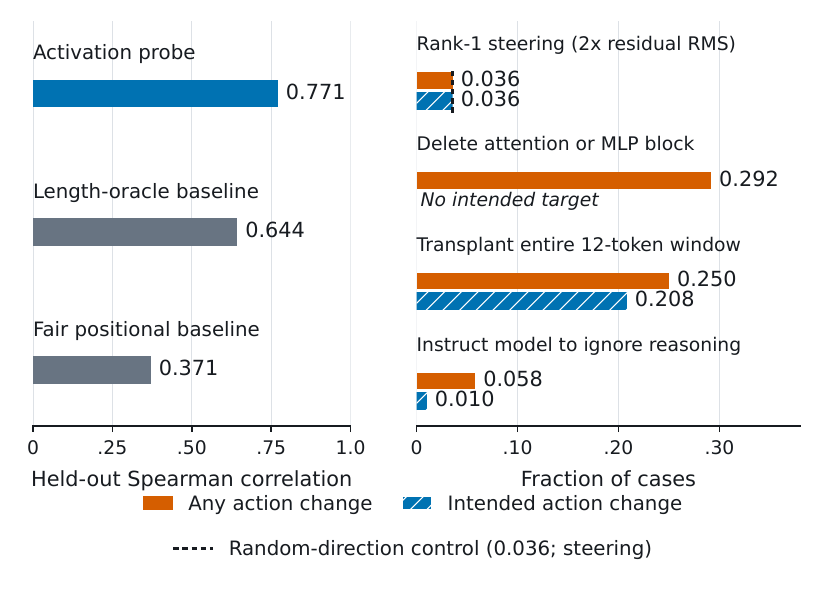}
\caption{
Decodability and intervention effectiveness at the reasoning-window site.
\textbf{Left:} decodability of action decisiveness against positional and length-based baselines.
\textbf{Right:} effectiveness of the tested activation interventions,
distinguishing any change in the forced action from changes to the intended target action.
}
\label{fig:readwrite}
\end{figure}
\subsection{Activation Patching}
\label{sec:activation-patching}
We next test whether action choice can be controlled through a reasoning-window representation from which action decisiveness is strongly decodable. Following activation-patching and activation-steering approaches \citep{zhang2024towards,turner2023steering}, we intervene on residual activations over a fixed window containing the final up to 12 reasoning tokens before the answer-forcing suffix.
From this site, a non-linear readout predicts inverse normalised action entropy with held-out Spearman $\rho=0.771\pm0.009$, compared with 0.644 for a length-oracle baseline and 0.371 for a positional baseline (\Cref{fig:readwrite}, left).

Targeted edits at this activation site are largely ineffective (\Cref{fig:readwrite}, right). 
Adding an action-associated rank-1 steering direction to the reasoning-window residual state \citep{turner2023steering} changes the action to the intended target in 3.6\% of cases, matching the 3.6\% rate for a norm-matched random direction. 
Replacing a low-dimensional subspace of the recipient representation with the corresponding donor representation produces no intended flips, while replacing the entire 12-token reasoning-window state redirects 20.8\% of donor--recipient pairs to the donor-associated action at the best tested layer. In contrast to the strong effects of reasoning-text interventions, action decisiveness is readily decodable at this site, while the tested activation edits provide limited control over action identity. These results do not show that the represented information is causally unused, only that the tested interventions do not reliably control the action. This is consistent with prior work showing that localisation need not identify
effective targets for behavioural editing \citep{hase2023does}.  Full intervention sweeps and controls are reported in Appendix~\ref{app:activation-interventions}.


\section{Discussion \& Limitations}
\label{sec:discussion}

Our results suggest that suboptimal behaviour in language model agents cannot be explained solely by whether task-relevant state information is represented correctly. In \Cref{sec:analysis1}, state information remains recoverable from model activations, yet the belief--action gap decreases after reasoning even as the decoded state distribution becomes less precise. Improved action consistency therefore does not appear to arise simply from forming a sharper estimate of the environment.
The report-based analyses show that explicit state information does not always guide action. In three illustrative wall-hit cases, the model correctly reports that the chosen direction is blocked and still selects it. Around commitment, report accuracy changes little and belief corrections rarely coincide with action changes (\Cref{sec:belief-commitment}). Action preferences sharpen gradually and usually stabilise around the middle of reasoning (\Cref{sec:reasoning}). Before commitment, preferences may change repeatedly, but we find no consistent correction point linking belief updates to improved action choice.
The causal interventions in \Cref{sec:causal} sharpen the distinction between information availability and causal control. Altering the reasoning text strongly changes the model's current action preference, whether by replacing a reasoning prefix or injecting a single action suggestion. By contrast, targeted edits to a reasoning-window representation from which action decisiveness is highly decodable do not reliably steer behaviour under the interventions tested. Thus, decodability does not imply causal control \citep{belinkov2022probing,hase2023does}. 
Our findings remain limited to a controlled deterministic grid world and to the estimators and interventions used here. Probe-based beliefs establish decodability rather than direct access to the model's internal beliefs \citep{hewitt2019designing,belinkov2022probing}, and the one-step IRL model recovers an inferred preference structure rather than the model's true value function. Likewise, the negative activation interventions apply only to the tested layers, subspaces, and edit families. Further work should test whether the observed dissociation between representation, commitment, and behavioural control persists in richer environments and under stronger causal interventions.


\section{Conclusion}
\label{sec:conclusion}

We study how language model agents transform represented state information into action by jointly analysing beliefs, inferred preferences, reasoning dynamics, and causal interventions in a controlled grid world. We find that task-relevant information can remain available even when behaviour is suboptimal, and that the belief--action gap decreases after reasoning despite less precise decoded state beliefs. During reasoning, action preferences gradually sharpen and stabilise well before the trace ends, while changes in explicit belief correctness explain only a small fraction of action revisions. 
Causal interventions further show that reasoning-text edits strongly redirect action preferences, whereas edits to a hidden representation predictive of action decisiveness are much less effective. 
Together, these results support treating state estimation, preference, and action selection as distinct sources of agent failure, and show that information being represented or decodable does not by itself imply that it governs behaviour.

\subsection*{AI use statement}

We used large language models to assist with code development and debugging, and editing parts of the manuscript. All AI-assisted code, and written content were reviewed by the authors before inclusion. Experimental results were obtained from the implemented analysis pipelines rather than generated by AI tools. We take responsibility for the final content of the paper, including all AI-assisted text, code, claims, and analyses.

\subsection*{Ethics Statement}

This work studies decision making in language model agents using a controlled grid-world environment and does not involve human participants or personal data. Our experiments are designed to characterize how task-relevant information is represented, reported, and translated into actions, rather than to improve a model's ability to manipulate or influence users.
The intervention methods in this work modify reasoning text or internal activations to test their causal effect on action selection. Although we use these methods only as analysis tools in a simple research environment, related techniques could in principle be used to steer model behaviour in less controlled settings. Our results should therefore be interpreted as evidence about the mechanisms studied here.
More broadly, we use terms such as belief, value, preference, and commitment as operational descriptions of measured model behaviour and internal representations. They should not be interpreted as evidence of human-like cognition, subjective experience, or intentionality.

\subsection*{Reproducibility statement}

We provide the experimental setup, state and belief definitions, value-estimation procedure, and main analysis methods in the paper. The appendix gives additional details on probe training, inverse reinforcement learning, reasoning segmentation, action commitment, report-based analyses, text interventions, activation interventions, controls, evaluation splits, hyperparameters, and robustness analyses.
Code for reproducing the experiments is provided through the repository linked in the paper. The repository contains the analysis pipelines used to obtain the reported results. Analyses operate on externally supplied trajectories, activations, model responses, and intervention outcomes, with the required inputs specified for each command.

\bibliography{ref}
\bibliographystyle{iclr2027_conference}

\appendix

\section{Additional Belief and Value Estimation Details}
\label{app:belief-value}

\subsection{Representation-Based Belief Estimation}
\label{app:representation-beliefs}

For each decision step $t$, token region $p$, and model layer $\ell$, we
construct an activation representation $\mathbf{u}_{t}^{(\ell,p)}$ by
concatenating the hidden states of the three tokens in the corresponding
readout region. We consider the three final prompt tokens before reasoning
and the three final reasoning tokens immediately before action generation.
Each token activation has 2,880 dimensions, giving an 8,640-dimensional
probe input.

We train separate probes for position, key possession, and door status.
Position is represented as a categorical variable over walkable grid
locations and is decoded using an MLP classifier. Key possession and door
status are binary variables and are decoded using linear classifiers. All
probes use the same state-disjoint split, with 80\% of unique task states
assigned to training and 20\% to evaluation. All occurrences of a given
state therefore appear in only one partition. Activation features are
standardised using statistics computed from the training split.

Let
$q^{\mathrm{probe}}_{t,p,\mathrm{pos}}(x,y)$,
$q^{\mathrm{probe}}_{t,p,\mathrm{key}}(h)$, and
$q^{\mathrm{probe}}_{t,p,\mathrm{door}}(d)$ denote the three predicted
component distributions. We combine them into a factorised distribution
over complete task states,
\begin{align}
    \widetilde q_{t,p}(s)
    &=
    q^{\mathrm{probe}}_{t,p,\mathrm{pos}}(x,y)
    q^{\mathrm{probe}}_{t,p,\mathrm{key}}(h)
    q^{\mathrm{probe}}_{t,p,\mathrm{door}}(d),
    \qquad
    s=(x,y,h,d),
    \label{eq:appendix-factorised-belief}
\end{align}
and renormalise over the valid state set $\mathcal S_G$,
\begin{align}
    \widehat q_{t,p}(s)
    &=
    \frac{
        \mathbbm{1}[s\in\mathcal S_G]\,
        \widetilde q_{t,p}(s)
    }{
        \sum_{s'\in\mathcal S_G}
        \widetilde q_{t,p}(s')
    }.
    \label{eq:appendix-joint-belief}
\end{align}

Belief uncertainty is measured using Shannon entropy,
\begin{align}
    H(\widehat q_{t,p})
    &=
    -\sum_{s\in\mathcal S_G}
    \widehat q_{t,p}(s)
    \log \widehat q_{t,p}(s).
    \label{eq:appendix-belief-entropy}
\end{align}

Expanded held-out decoding results are reported in
Appendix~\ref{app:probe-results}.

\subsection{Report-Based Belief Estimation}
\label{app:report-beliefs}

We obtain report-based beliefs by directly querying the model about features
of the same rendered environment state supplied during the original decision.
The full query registry contains 38 questions comprising 30 categorical
questions and eight coordinate questions.

The categorical queries ask whether the model currently holds the key,
whether the door is open, whether the cell immediately in each cardinal
direction is a wall, whether the goal, key, or door is adjacent in each
direction, and what would happen after each candidate movement. The latter
queries test whether the movement would hit a wall, leave the agent holding
the key, or leave the door open. Coordinate queries ask for the current
locations of the agent, goal, key, and door and for the successor location
under each candidate action.

Categorical prompts request one of
$\{\texttt{yes},\texttt{no},\texttt{unknown}\}$.
For repeated generations, the report distribution is the empirical frequency
of each parsed answer. When candidate scores are available, we normalise their
logits,
\begin{align}
    q^{\mathrm{report}}_{t,p,j}(z)
    &=
    \frac{
        \exp(\ell_{t,p,j}(z))
    }{
        \sum_{z'\in\mathcal Z_j}
        \exp(\ell_{t,p,j}(z'))
    }.
    \label{eq:appendix-report-belief}
\end{align}
Responses that do not match the expected categorical or coordinate format are
recorded as invalid rather than matched approximately. Report accuracy is
computed from the most probable parsed answer, and uncertainty is measured
using Shannon entropy.

Because report queries require a separate forward pass, they may elicit
information that was not used in the original decision. We therefore analyse
them separately rather than incorporating them into the belief--action gap.

Results from the failure-focused report analyses are provided in
Appendix~\ref{app:report-results}.

We additionally examine a failure-focused set containing three decisions in
which the emitted action points towards a blocked neighbouring cell. In all
three cases, the model correctly reports that the selected direction is
blocked while nevertheless choosing that action. These examples provide a
direct case in which task-relevant state information is explicitly
recoverable but does not determine behaviour.

\subsection{One-Step IRL Estimation}
\label{app:irl-details}

For each reasoning position $p$, we represent an environment state $s$ using
the mean activation representation across occurrences of that state,
\begin{align}
    \boldsymbol{\phi}_p(s)
    &=
    \frac{1}{|\mathcal I_p(s)|}
    \sum_{t\in\mathcal I_p(s)}
    \mathbf{u}_{t}^{(p)},
    \qquad
    \mathcal I_p(s)=\{t:s_t=s\}.
    \label{eq:appendix-state-representation}
\end{align}
Pre-reasoning representations are constructed from the prompt-suffix
activations of the no-reasoning condition, while post-reasoning
representations use the final reasoning-window activations. The
representations are standardised separately before fitting the corresponding
value models.

The value function is linear,
\begin{align}
    g_{\theta_p}(\boldsymbol{\phi}_p(s))
    &=
    \theta_p^\top\boldsymbol{\phi}_p(s),
\end{align}
and actions are evaluated through their deterministic successor states. The
resulting Boltzmann policy is
\begin{align}
    p_{\theta_p}(a\mid s,G)
    &=
    \frac{
        \exp\!\left[
            \beta g_{\theta_p}
            \bigl(\boldsymbol{\phi}_p(T_G(s,a))\bigr)
        \right]
    }{
        \sum_{a'\in\mathcal A}
        \exp\!\left[
            \beta g_{\theta_p}
            \bigl(\boldsymbol{\phi}_p(T_G(s,a'))\bigr)
        \right]
    }.
    \label{eq:appendix-irl-policy}
\end{align}

Not every successor state encountered in a one-step comparison is necessarily
represented in the trajectory data. We therefore retain a source-state
observation only when activation-derived representations are available for
the successors of all four candidate actions. This ensures that every
training observation defines the complete four-action comparison appearing
in \Cref{eq:appendix-irl-policy}.

For the agent-supervised model, the pre-reasoning value function is trained
against the action produced directly from the environment prompt without
reasoning, while the post-reasoning value function is trained against the
action emitted after reasoning. This ensures that the representation and
action labels correspond to the same decision condition.

We fit the models by minimising
\begin{align}
    \mathcal L(\theta_p)
    &=
    -\sum_{(G,s,a)\in\mathcal D_p}
    \log p_{\theta_p}(a\mid s,G).
    \label{eq:appendix-irl-objective}
\end{align}
The optimisation uses Adam for 500 epochs with an initial learning rate of
$10^{-3}$, cosine learning-rate annealing, gradient clipping at norm $1.0$,
and $L_2$ regularisation. Source states are split into disjoint training and
validation sets so that every occurrence of a source state belongs to only
one partition. Unless otherwise stated, we use $\beta=1$.


Robustness to the inverse-temperature parameter is reported in
Appendix~\ref{app:irl-results}.

\subsection{Planner-Supervised IRL Control}
\label{app:planner-irl-methods}
As a control, we replace the agent's realised action labels with
planner-optimal actions while keeping the activation representations and value
model otherwise unchanged. For each state $s$, the planner computes the set
of shortest-path optimal actions
\begin{align}
    A_G^*(s)
    &=
    \underset{a\in\mathcal A}{\operatorname{argmin}}\;
    D_G^*(T_G(s,a)).
\end{align}
A planner-optimal action is used as the supervision target during fitting, while evaluation can additionally treat any action in $A_G^*(s)$ as correct when multiple shortest-path actions are available.

This control asks whether the activation-derived state representations support a task-optimal action ordering more readily than they reproduce the agent's realised policy. 


\section{Report-Based Probe Protocol}
\label{app:report-protocol}

For report-based belief estimation, we use a total of 38 questions: 30 categorical questions and eight coordinate questions.
Every prompt begins by describing the fully observable $9\times9$ DoorKey environment, giving the symbol legend, and defining \textsc{right}, \textsc{left}, \textsc{up}, and \textsc{down} as cardinal directions in the rendered grid.
It then supplies the current grid.
Endpoint prompts include the carrying-key status for questions configured to require it; reasoning-time prompts include this status and the reasoning prefix $R_{t,\leq p}$.

\paragraph{Categorical question templates.}
The 30 categorical questions are generated from the following templates, where $d\in\{\textsc{right},\textsc{left},\textsc{up},\textsc{down}\}$, $a$ ranges over the corresponding four moves, and $o\in\{\text{goal},\text{key},\text{door}\}$:
\begin{enumerate}
    \item ``Given the current grid only, do you currently have the key?'' and ``Given the current grid only, is any door currently open?'' (two questions);
    \item ``Given the current grid only, is the cell immediately $d$ of the agent a wall (\texttt{\#})?'' (four questions);
    \item ``Given the current grid only, is the $o$ immediately to your $d$?'' (12 questions);
    \item ``Given the current grid only, if you move $a$ now, will you hit a wall?'', ``will you have the key after that move?'', and ``will the grid still show an open door (\texttt{O}) after that move?'' (12 questions).
\end{enumerate}
The requested output is exactly one label: \texttt{A=yes}, \texttt{B=no}, or \texttt{C=unknown}.

\paragraph{Coordinate question templates.}
Four questions ask for the current location of the agent, goal, key, or a door, and four ask for the agent's location after each candidate move.
The requested output is exactly \texttt{\{"row": <int>, "col": <int>\}}.
For a current object that is not visible, the prompt specifies $(-1,-1)$; rows and columns are zero-indexed grid coordinates.

\paragraph{Post-processing and scores.}
Categorical text is lower-cased and stripped of surrounding quotation marks and terminal punctuation.
Only an exact \texttt{A}/\texttt{B}/\texttt{C} or \texttt{yes}/\texttt{no}/\texttt{unknown} is accepted; other text is \texttt{invalid} rather than being matched fuzzily.
For repeated generations, every sample, including \texttt{invalid}, contributes to the empirical distribution.
For API log-probability runs, we exponentiate and renormalise the recognised \texttt{A}/\texttt{B}/\texttt{C} mass at the answer-token position and record missing candidates and excluded top-$k$ mass.
For the local scaled run, we score all three candidate tokens directly at the final answer position, divide the logits by temperature $0.7$, and apply softmax.

For coordinate responses, we extract a JSON object and accept it only when both \texttt{row} and \texttt{col} are integers.
Repeated coordinate samples are summarised by their modal valid coordinate; we additionally record parse rate, the number of distinct coordinates, and mean Manhattan error.

\paragraph{Panels used.}
The reasoning-time panel uses \texttt{wall\_left/right/up/down}, \texttt{has\_key}, and \texttt{door\_open}, plus \texttt{hit\_wall\_after\_$a$}, \texttt{has\_key\_after\_$a$}, and \texttt{door\_open\_after\_$a$} for the action read out at that position.
Earlier small-scale reasoning runs additionally query \texttt{agent\_location}, \texttt{goal\_location}, \texttt{key\_location}, and \texttt{door\_location}; the scaled matched-cohort run defers these coordinate questions.
Other endpoint analyses select question families from the same 38-question registry.

\begin{table}[htbp]
\centering
\small
\begin{tabular}{lccc}
\toprule
Query family & State--question pairs & Greedy & Ten-sample modal \tabularnewline
\midrule
Wall & 132 & 93.9\% & 97.0\% \tabularnewline
Goal adjacency & 132 & 100.0\% & 100.0\% \tabularnewline
Door adjacency & 132 & 99.2\% & 100.0\% \tabularnewline
Key adjacency & 132 & 67.4\% & 97.0\% \tabularnewline
\bottomrule
\end{tabular}
\caption{Report accuracy on the 33-state failure-mode set. Each family contains four directional questions per state. The modal result uses ten samples per question at temperature 0.7.}
\label{tab:report-accuracy-33}
\end{table}

\subsection{Additional Report-Based Belief Results}
\label{app:report-results}

We use the report-based estimates to test whether task-relevant information
that can be explicitly recovered from the model is reflected in its behaviour.
On the 33-state failure set, each directional query family contains 132
state--question pairs. Greedy accuracy is 93.9\% for walls, 100.0\% for goal
adjacency, 99.2\% for door adjacency, and 67.4\% for key adjacency. Taking
the modal answer from ten samples at temperature 0.7 increases these values to
97.0\%, 100.0\%, 100.0\%, and 97.0\%, respectively
(\Cref{tab:report-accuracy-33}).

We additionally analyse a 13-decision failure-focused slice containing four avoidable detours, three wall-hit decisions, two backtracks, and four optimal baseline decisions. A wall-hit decision is one in which the emitted action points towards a blocked neighbouring cell and therefore fails to move the agent in the intended direction. Reports are elicited from the same rendered
environment while varying the fraction of the original reasoning trace shown to the model, without revealing the final action. In all three wall-hit failures, the model correctly reports that the direction ultimately selected by the agent is blocked. These examples illustrate cases in which the local information needed to avoid the error is explicitly recoverable despite the model selecting the blocked action.

\paragraph{Comparison with representation-based estimates.}
We also compare report-based and representation-based estimates on a matched
set of 15 decision states. For each state, we ask whether the adjacent cell in
each cardinal direction is blocked. Report-based estimates identify the wall
status correctly for every evaluated direction, whereas decoding from the
original decision activations is less reliable
(\Cref{tab:matched-wall-app}).

\begin{table}[htbp]
\centering
\small
\begin{tabular}{lccc}
\toprule
\textbf{Direction} &
\textbf{Report-based} &
\textbf{Probe (Pre-reasoning)} &
\textbf{Probe (Post-reasoning)} \\
\midrule
\textsc{left}  & 100.0\% & 73.3\%  & 73.3\% \\
\textsc{right} & 100.0\% & 86.7\%  & 93.3\% \\
\textsc{up}    & 100.0\% & 100.0\% & 93.3\% \\
\textsc{down}  & 100.0\% & 80.0\%  & 93.3\% \\
\bottomrule
\end{tabular}
\caption{
Accuracy of report-based and representation-based estimates of whether the
adjacent cell in each direction is blocked, evaluated on the same 15 decision
states.
}
\label{tab:matched-wall-app}
\end{table}

The comparison should not be interpreted as showing that report-based beliefs
are intrinsically more faithful than activation-based estimates. Reports are
elicited in a separate forward pass and may recover information that was not
used during the original decision. Rather, they show that relevant local state
information can sometimes be explicitly retrieved even when it is not
perfectly decoded from the original decision activations or reflected in the
selected action.


\section{Additional Belief and Value Results}
\label{app:belief-value-results}

\subsection{Expanded State-Decoding Results}
\label{app:probe-results}

The main text reports a compact set of metrics for representation-based state
decoding. Here we provide the full held-out evaluation. Probe construction,
training, and the state-disjoint split are described in
Appendix~\ref{app:representation-beliefs}.

Key possession and door status are decoded reliably before reasoning and less
accurately after reasoning. Exact position is substantially harder to recover.
Across the 39 walkable grid locations, uniform top-1 chance accuracy is
2.56\%. The position probe reaches 34.8\% exact accuracy before reasoning and
13.1\% after reasoning, remaining above this chance level in both conditions.

Although exact position decoding is limited, the decoded distribution is
usually spatially close to the true location, particularly before reasoning.
Before reasoning, the most probable decoded location is within Manhattan
distance one of the true position in 95.5\% of held-out examples and within
distance two in 99.0\%. The corresponding values after reasoning are 60.8\%
and 83.5\%. The same pattern appears when considering the full probability
distribution rather than only its most probable location.

\begin{table}[htbp]
\centering
\small
\begin{tabular}{lcc}
\toprule
\textbf{Metric} & \textbf{Pre-reasoning} & \textbf{Post-reasoning} \\
\midrule
\texttt{has\_key} accuracy & 100.0\% & 89.5\% \\
\texttt{door\_open} accuracy & 100.0\% & 88.6\% \\
Location top-1 accuracy & 34.8\% & 13.1\% \\
Top-1 within Manhattan distance 1 & 95.5\% & 60.8\% \\
Top-1 within Manhattan distance 2 & 99.0\% & 83.5\% \\
Mean probability mass within distance 1 & 93.1\% & 59.2\% \\
Mean probability mass within distance 2 & 97.2\% & 81.5\% \\
\bottomrule
\end{tabular}
\caption{
Expanded representation-based state decoding on held-out environment states.
Location top-1 accuracy requires an exact match. ``Top-1 within Manhattan
distance $k$'' measures whether the most probable decoded location lies within
distance $k$ of the true location. ``Mean probability mass within distance
$k$'' is the probability assigned to all locations within that distance of the
true position.
}
\label{tab:rep-beliefs-full}
\end{table}

The decline after reasoning is therefore not restricted to exact location
classification. Both local probability mass and binary state-variable
decoding also become less precise at the post-reasoning readout.

\subsection{Planner-Supervised IRL Control}
\label{app:planner-irl-results}
The main analysis fits the one-step value model to the actions realised by the
language model. As a control, we fit the same model using planner-optimal
actions as supervision while leaving the activation-derived state
representations and model class unchanged. For each source state $s$, the
planner computes the set of shortest-path optimal actions $A_G^*(s)$. When
several actions are tied, one is used as the training label, while evaluation
can additionally count any member of $A_G^*(s)$ as optimal.

The purpose of this control is to distinguish two questions. Agent-supervised
IRL asks how well the activation-derived state representation supports the
model's realised policy, whereas planner-supervised IRL asks how well the same
representation supports the task-optimal action structure. The latter is
substantially more recoverable, indicating that the state representations
contain a cleaner task-optimal ordering than is expressed in the agent's
realised actions.

A separate four-room label-control experiment provides the same comparison
while holding the activation inputs and training procedure fixed. Replacing
agent-action labels with A$^\star$-optimal labels increases test accuracy from
45.4\% to 96.0\%, and optimal-set membership accuracy from 36.9\% to 99.2\%.
This separation therefore persists when the supervision labels are the main
quantity changed.

\begin{figure}[htbp]
    \centering
    \includegraphics[width=0.8\textwidth]{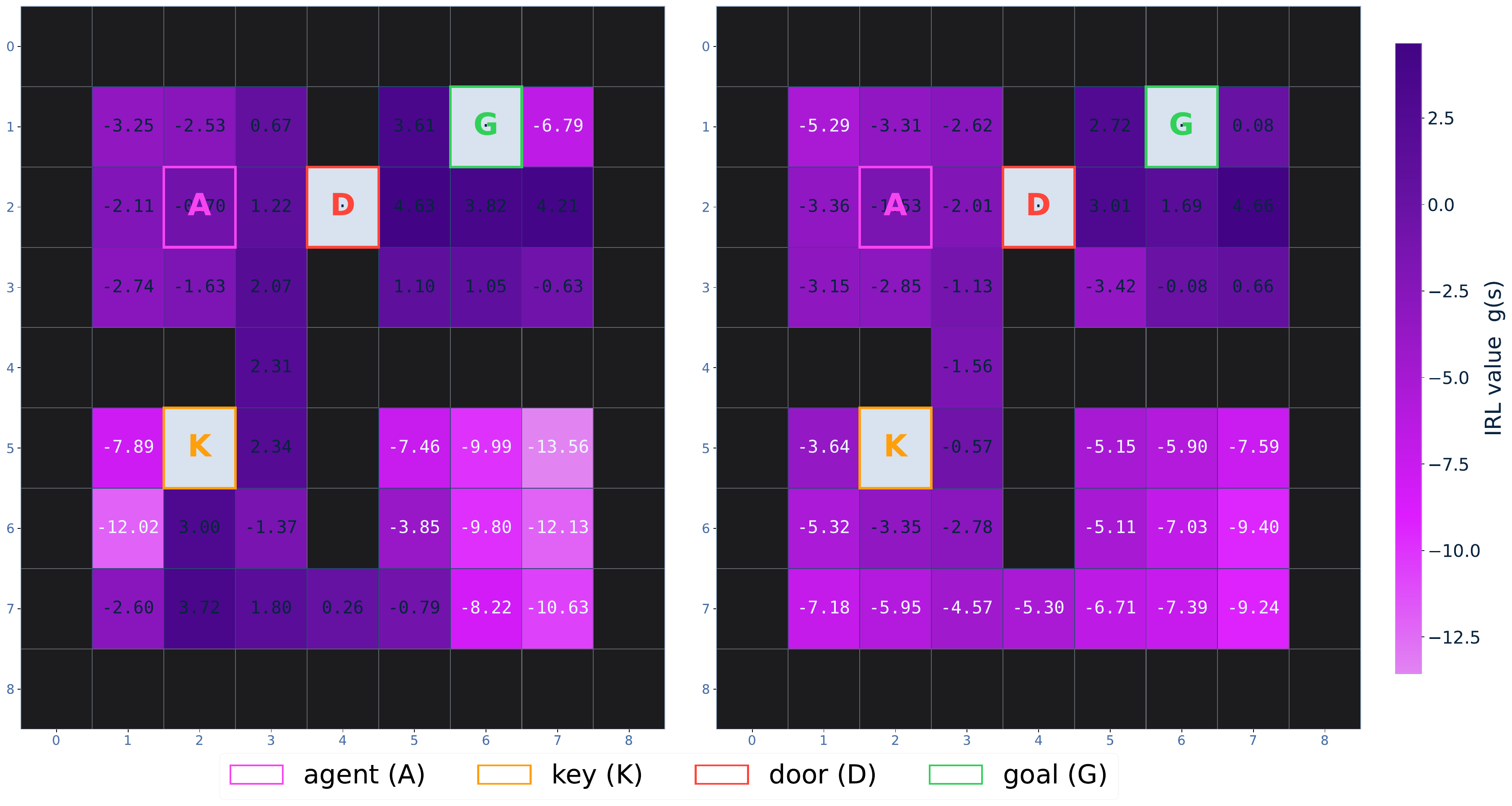}
    \caption{
    Example learned state values for planner-supervised IRL (left) and
    agent-supervised IRL (right), using post-reasoning activation-derived
    state representations. Darker cells indicate larger learned values.
    Because the one-step objective identifies relative successor-state
    preferences, absolute values should not be interpreted as goal distances;
    the comparison is intended to visualise differences in the learned
    state-preference structure.
    }
    \label{fig:irl-value-app}
\end{figure}

\Cref{fig:irl-value-app} visualises the learned state values for
planner-supervised and agent-supervised models for one task phase. The
absolute magnitudes and signs should not be interpreted as distances to the
goal. One-step IRL constrains values only through comparisons among successor
states, so the meaningful quantity is the induced ordering of candidate
actions rather than the absolute value assigned to an individual cell.

\subsection{Agent-Supervised IRL and Inverse-Temperature Robustness}
\label{app:irl-results}

The agent-supervised value models are fitted separately for the pre-reasoning and post-reasoning conditions. The pre-reasoning model is supervised by the action elicited directly from the environment prompt without intervening reasoning, while the post-reasoning model is supervised by the action emitted after the reasoning trace. This aligns the action labels with the corresponding activation representations.

On held-out source states, the resulting model predicts 29.2\% of
pre-reasoning actions and 45.2\% of post-reasoning actions. The lower
pre-reasoning accuracy coincides with a substantially more imbalanced
no-reasoning action distribution, making direct comparison of the two raw accuracies less informative than their use as condition-specific preference models in the belief--action gap analysis.

We additionally repeat the IRL and belief--action gap analyses across several values of the inverse-temperature parameter $\beta$. Changing $\beta$ alters the sharpness of the induced Boltzmann policy and the numerical scale of the learned values, but does not change the qualitative endpoint result. Across the tested settings, the belief--action gap decreases from before to after reasoning while decoded belief entropy increases. We therefore use $\beta=1$ for the main analysis.


\section{Additional Reasoning-Time Analyses}
\label{app:reasoning-time}

\subsection{Reasoning Segmentation and Commitment Measurement}
\label{app:commitment-methods}

For the reasoning-time analysis, we use 95 trajectories comprising 1,276
decisions and 2,370,850 reasoning tokens. The median reasoning length is
1,480 tokens per decision, with an interquartile range of 642--2,605 tokens
and a maximum of 9,733.

We segment each reasoning trace into sentence-level units before aligning
behavioural and activation measurements. A terminal JSON action of the form
\texttt{\{"action": "DIRECTION"\}} is first removed, while embedded action
JSON remains part of the reasoning trace. Sentence boundaries are introduced
at blank lines and after periods, question marks, or exclamation marks followed
by whitespace. Boundaries are not introduced after periods occurring in
decimal numbers, coordinates, common abbreviations such as \texttt{e.g.},
\texttt{i.e.}, \texttt{etc.}, and \texttt{vs.}, or numbered action entries
such as \texttt{1. LEFT}. Empty spans are discarded, and spans containing
fewer than ten non-whitespace characters are merged with a neighbouring span.

We then tokenize the complete reasoning trace using the
\texttt{gpt-oss-20b} tokenizer and assign each token to the sentence whose
character span it overlaps. This yields 153,622 sentence-level reasoning units.
The segmentation is fixed independently of the intermediate prefixes used for
action elicitation.

For action-preference measurements, we evaluate at most 80 prefixes per decision. Consecutive prefixes are typically separated by 20 tokens, with wider spacing for longer reasoning traces, yielding 98,310 evaluated prefixes in total. At each prefix, we append a fixed answer-forcing suffix and obtain the logits for
$\{\textsc{left},\textsc{right},\textsc{up},\textsc{down}\}$.
Applying a softmax gives the prefix-level action distribution
$\pi_{t,k}(a)$.

The preferred action at prefix $k$ is
\begin{align}
    \hat a_{t,k}
    &=
    \arg\max_{a\in\mathcal A}\pi_{t,k}(a).
\end{align}
We define action commitment as the earliest evaluated prefix after which this
preferred action never changes again,
\begin{align}
    k_t^{\mathrm{com}}
    &=
    \min
    \left\{
        k:
        \hat a_{t,k'}=\hat a_{t,k}
        \quad
        \forall k'\geq k
    \right\}.
    \label{eq:appendix-commitment}
\end{align}
This is therefore the start of the final stable segment of the preferred-action
sequence and is defined for all 1,276 decisions.

We additionally identify the first decisional mention of an action in the reasoning text. A decisional mention is an occurrence of
\textsc{up}, \textsc{down}, \textsc{left}, or \textsc{right} used as a candidate move, including bare direction words, mentions following a movement verb or answer cue, and capitalised answer-like uses. Descriptive uses such as ``to the left of the door'', idioms such as ``pick up the key'', and negated mentions such as ``cannot go up'' are excluded using fixed context rules.

Textual commitment is defined separately from forced-action commitment. It is the first sentence classified as concluding or answering that explicitly mentions the action eventually emitted by the model and is not followed by a sentence mentioning a different action. A textual commitment is identified in 753 of the 1,276 decisions.

\subsection{Additional Commitment Results}
\label{app:commitment-results}

The prompt alone induces a non-uniform prior over actions. At the empty prefix,
the mean probability assigned to the action eventually emitted is 0.34,
compared with 0.25 under a uniform four-action distribution, and mean inverse
normalised action entropy is 0.30. The initial preference is strongest for
decisions that eventually emit \textsc{right}, where the mean initial
probability is 0.66, compared with 0.37 for \textsc{left} and 0.14 for both
\textsc{up} and \textsc{down}.

To distinguish confidence in the eventual action from overall decisiveness, we
track both
$c_{t,k}=\pi_{t,k}(a_t)$ and inverse normalised entropy,
\begin{align}
    1-\frac{H(\pi_{t,k})}{\log 4}.
\end{align}
Both increase throughout reasoning, as reported in the main text.

Commitment typically occurs around the middle of the reasoning trace. The
median position of the first decisional action mention is 0.30 of the trace,
the median action-commitment position is 0.49, and the median position after
which $c_{t,k}$ permanently exceeds 0.8 is 0.65. Only 37.0\% of decisions
commit before the first decisional action mention.

A substantial portion of reasoning therefore follows action commitment. The
median post-commitment share is 50.6\% of the reasoning trace, corresponding
to a median of 524 tokens and a mean of 887 tokens. The relative
post-commitment segment is larger for shorter traces: median commitment
position is 0.32 in the shortest quartile of reasoning length and 0.60 in the
longest quartile. Because commitment is defined only over evaluated prefixes,
the inferred commitment position is an upper-resolution estimate of the true
stabilisation time.

We also examine how stable the current preference is before retrospective
commitment. The probability that the preferred action changes at the next
evaluated prefix falls steadily from 0.133 in the first decile of reasoning to
0.019 in the final decile. Decisions undergo 6.0 preference switches on
average before commitment. Current confidence predicts, but does not guarantee,
future stability: when $c_{t,k}>0.9$, the probability of at least one later
switch is still 0.089.

Commitment timing also differs across eventual actions. Decisions ending in
\textsc{right} commit earliest, at a median normalised position of 0.39,
whereas those ending in \textsc{down} commit latest, at 0.61.

\begin{figure}[t]
\centering
\begin{subfigure}[b]{0.42\textwidth}
    \centering
    \includegraphics[width=\linewidth]{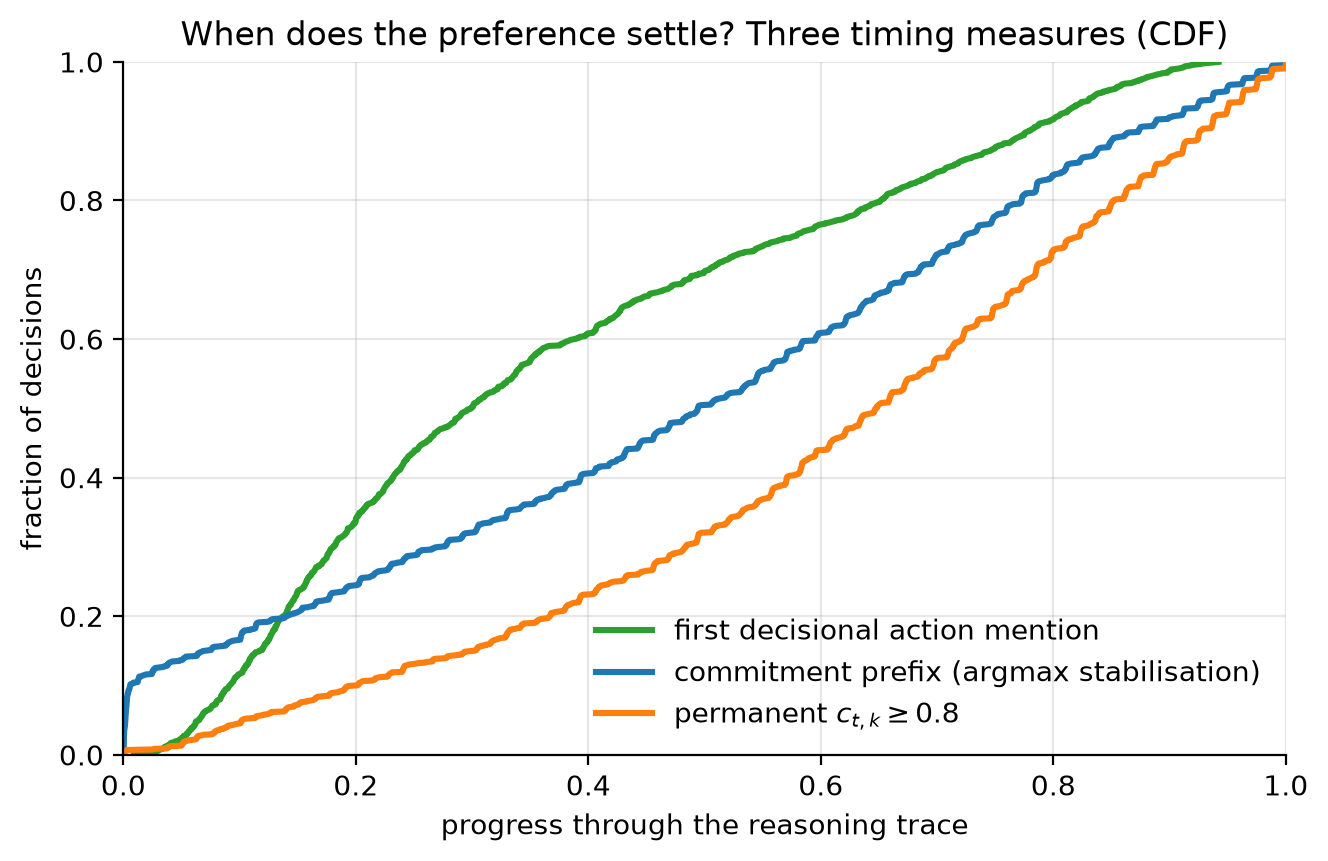}
    \caption{}
    \label{fig:three-clocks-app}
\end{subfigure}
\hfill
\begin{subfigure}[b]{0.56\textwidth}
    \centering
    \includegraphics[width=\linewidth]{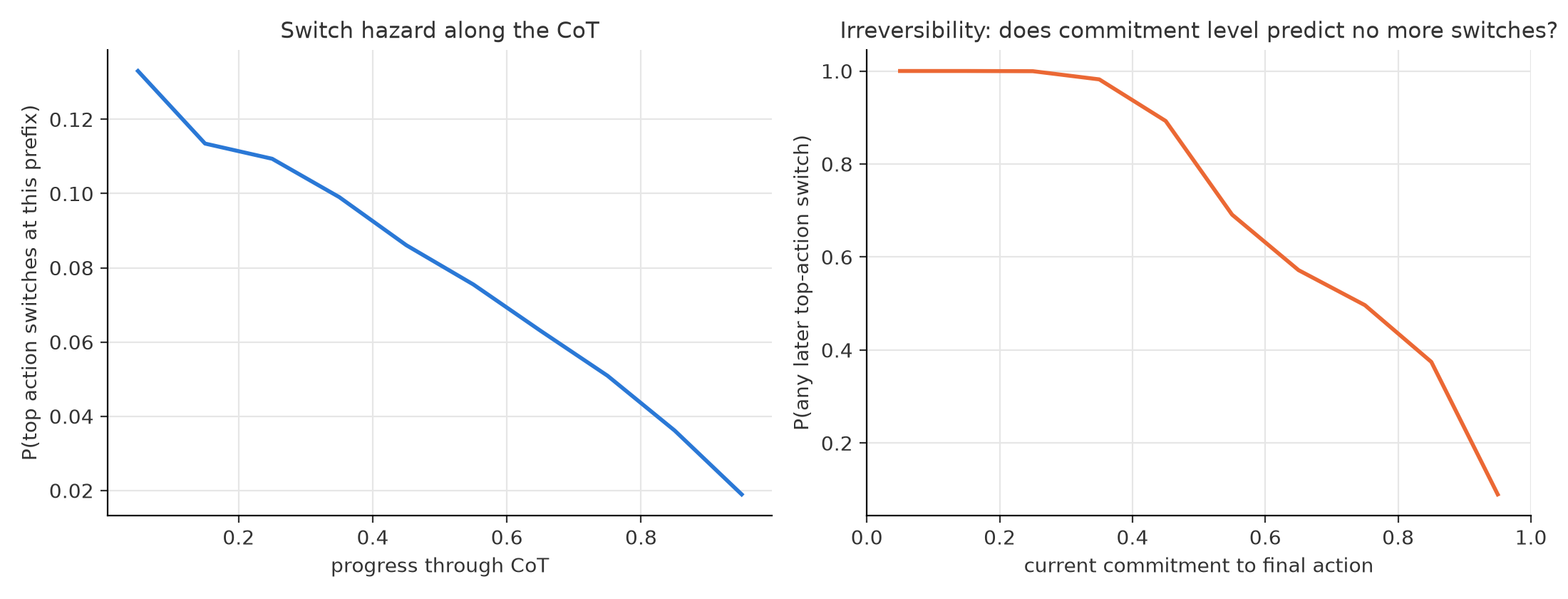}
    \caption{}
    \label{fig:switch-hazard-app}
\end{subfigure}
\caption{
Additional commitment timing and stability results.
\textbf{(a)} Cumulative distributions of the first decisional action mention,
the action-commitment prefix, and the point after which the eventual-action
probability remains above 0.8.
\textbf{(b)} Left: probability that the preferred action changes at the next
evaluated prefix as reasoning progresses. Right: probability of at least one
later preference switch as a function of the current probability assigned to
the eventual action.
}
\label{fig:commitment-dynamics-app}
\end{figure}

\paragraph{Reasoning content before and after commitment.}
For the 1,004 decisions with sentence-level function annotations, we classify
141,377 reasoning sentences into four coarse categories: world-reading,
solving, meta-checking, and concluding or answering. Labels are produced by
\texttt{gpt-oss-20b} using a fixed rubric adapted from
\citet{bogdan2025thought} and compared with 152 blind author-labelled
sentences. Precision is 0.91 for concluding or answering, 0.78 for solving,
0.73 for world-reading, and 0.32 for meta-checking; we therefore treat the
meta-checking category as exploratory.

Before commitment, 68\% of sentences are classified as world-reading, compared
with 29\% after commitment. Solving increases from 25\% to 45\%, while
concluding or answering increases from 1\% to 12\%. Against a placebo in which
the commitment position is replaced by a position sampled from traces of
similar length, the reduction in world-reading and increase in solving are
larger than expected from ordinary progress through the trace, whereas the
increase in concluding or answering is not.

A complementary keyword analysis gives the same broad pattern. After
commitment, grid-related vocabulary falls to 0.37 times its pre-commitment
density, coordinate pairs increase by a factor of 1.41, and checking-related
vocabulary increases by a factor of 3.73. Direction words used as explicit
answer mentions occur almost entirely after commitment, with 29 occurrences
before and 1,751 after.

\begin{figure}[t]
\centering
\includegraphics[width=\textwidth]{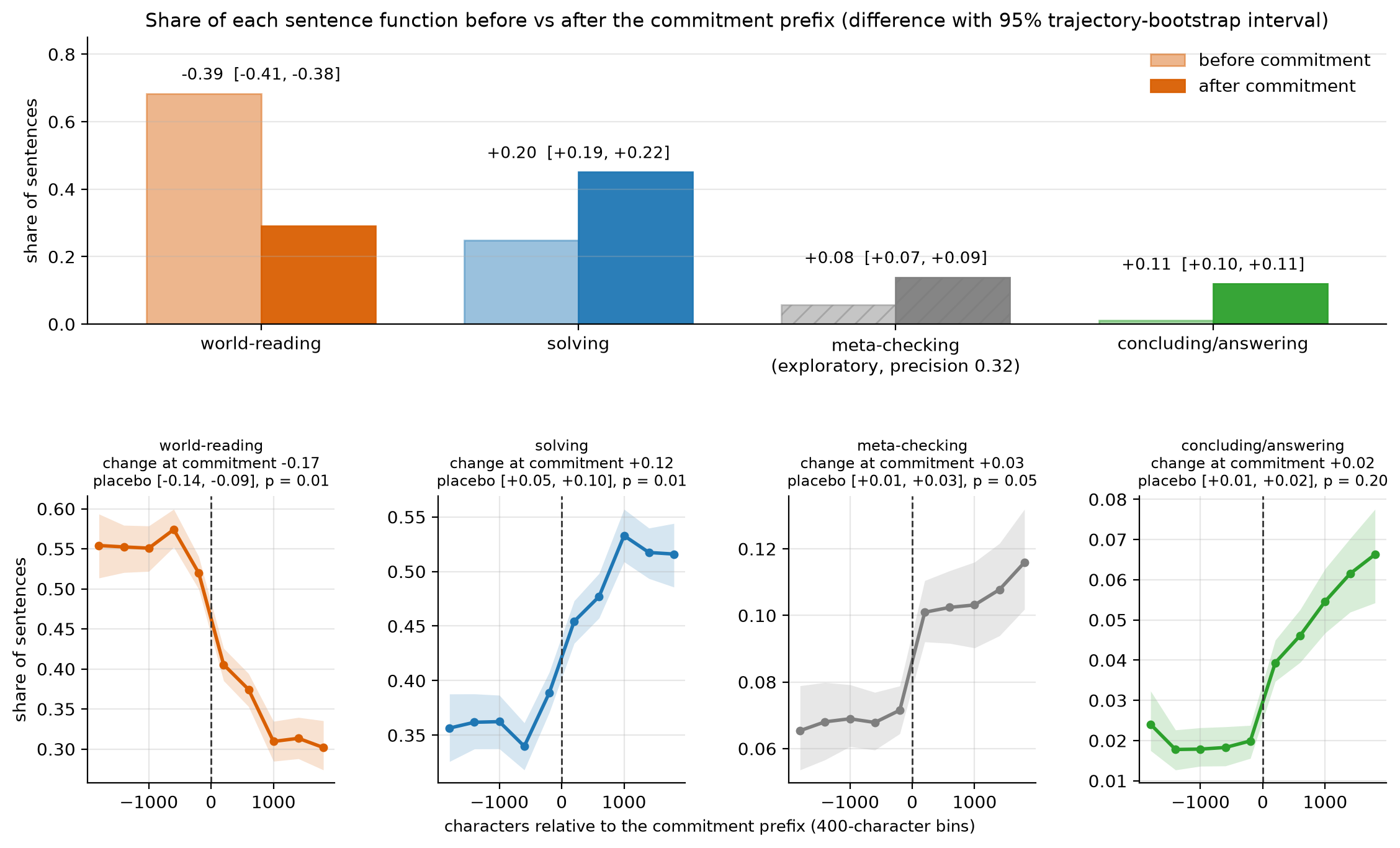}
\caption{
Sentence-function distributions around action commitment for 1,004 decisions
and 141,377 sentences. The top panel compares function frequencies before and
after commitment. The bottom panel shows their local evolution around the
commitment position and compares the observed change with a position-matched
placebo. Meta-checking should be interpreted as exploratory because its
annotation precision is low.
}
\label{fig:sentence-functions-app}
\end{figure}

\subsection{Belief Changes Around Commitment and Action Revision}
\label{app:belief-commitment}

The reasoning-time report dataset contains 46 matched decisions from 31
trajectories, covering 7,084 sentence prefixes and 42,504 current-state
reports. The reports comprise four local wall questions together with key
possession and door status. Of these decisions, 45 from 30 trajectories contain
reports both before and at or after the retrospective action-commitment point.

For each decision, we average report entropy and accuracy separately over
prefixes before commitment and over prefixes at or after commitment, and then
average these decision-level values across decisions. Mean report entropy
decreases from 0.392 to 0.336 bits, corresponding to a change of
$-0.056$ bits with a 95\% trajectory-bootstrap interval of
$[-0.105,-0.009]$. Mean report accuracy changes only from 82.8\% to 83.0\%,
an increase of 0.2 percentage points.

\begin{table}[t]
\centering
\small
\begin{tabular}{lrrr}
\toprule
Measure & Before & At or after & Change [95\% CI] \\
\midrule
Belief entropy (bits)
    & 0.392 & 0.336 & $-0.056\;[-0.105,-0.009]$ \\
Report accuracy
    & 82.8\% & 83.0\% & $+0.2\;[0.0,0.6]$ pp \\
\bottomrule
\end{tabular}
\caption{
Changes in report-based beliefs around retrospective action commitment.
Values are averaged within decisions before being aggregated across decisions.
Intervals are 95\% trajectory-clustered bootstrap intervals.
}
\label{tab:belief-commitment-changes-app}
\end{table}

We next test whether individual revisions of the preferred action coincide
with discrete changes in report correctness. Across 682 action-change events,
the correctness of at least one of the six current-state reports changes in
only 14 cases, or 2.1\%. For optimal-to-suboptimal changes, this occurs in 6
of 161 events, and for suboptimal-to-optimal changes in 3 of 182 events.

The corresponding change in mean report error is close to zero in all cases.
For all action changes, the mean difference is 0.0005 with a 95\% interval of
$[-0.0022,0.0027]$. For optimal-to-suboptimal changes it is 0.0000 with
interval $[-0.0065,0.0055]$, and for suboptimal-to-optimal changes it is
$-0.0018$ with interval $[-0.0080,0.0017]$.

\begin{table}[t]
\centering
\small
\begin{tabular}{lrrr}
\toprule
Transition & Events & Any report change & $\Delta$ error [95\% CI] \\
\midrule
Any action change
    & 682 & 14 (2.1\%) & $0.0005\;[-0.0022,0.0027]$ \\
Optimal to suboptimal
    & 161 & 6 (3.7\%) & $0.0000\;[-0.0065,0.0055]$ \\
Suboptimal to optimal
    & 182 & 3 (1.6\%) & $-0.0018\;[-0.0080,0.0017]$ \\
\bottomrule
\end{tabular}
\caption{
Changes in report correctness at action transitions. ``Any report change''
counts events where the correctness of at least one of the six current-state
reports changes. Error changes compare the preceding sentence with the
transition sentence.
}
\label{tab:belief-action-transitions-app}
\end{table}

Notably, all six current-state reports remain correct across 66 of the 161
optimal-to-suboptimal transitions. Thus, action quality can deteriorate even
without a corresponding error in the reported local state.

\paragraph{Prospective monitoring.}
We additionally test whether current behavioural and belief readouts predict an
optimality loss within the next three sentence boundaries. Among 5,283
prefixes at which the current preferred action is optimal, 347 are followed by
such a loss, corresponding to a random-ranking AUPRC of 0.066. A behavioural
baseline using reasoning progress, current action confidence, and the
Jensen--Shannon divergence from the preceding action distribution reaches an
AUPRC of 0.273 on held-out trajectories. Adding the observable belief reports
raises AUPRC to 0.334, an improvement of 0.061 with a 95\% trajectory-bootstrap
interval of $[0.014,0.123]$. The analogous next-sentence prediction is
inconclusive.

\paragraph{Attention at action changes.}
As an exploratory analysis, we compare attention from the immediate action
readout to the newest reasoning sentence at action-change events against a
progress-matched stable-action control from the same decision. At layer 15,
attention to the newest sentence increases by 0.0838 percentage points per
token at action changes across 67 matched pairs, with a 95\% interval of
$[0.0442,0.1314]$. Corresponding changes for the grid, key-status statement,
and earlier reasoning include zero. This association is observational and
does not establish that the attended sentence caused the action revision.

\subsection{Hidden-State Clustering Analysis}
\label{app:reasoning-clustering}

We represent each reasoning sentence using mean-pooled layer-15 residual
activations from \texttt{gpt-oss-20b}, yielding one 2,880-dimensional vector
per sentence. For clustering, we sample 25,000 of the 153,622 sentence
representations, reduce them to 100 principal components retaining 77.84\% of
the variance, and $\ell_2$-normalise the resulting vectors.

We use MiniBatch K-means with $K=12$ clusters for the main analysis.
Alternative clustering methods were also evaluated. HDBSCAN produces more
locally separated clusters but labels 77.1\% of sentence-mean representations
as noise. Sentence-mean K-means provides complete coverage and stronger
inter-cluster dispersion and Calinski--Harabasz score among methods that assign
every sentence to a cluster. We therefore use K-means for the main
reasoning-time analysis.

The clustering is entirely unsupervised. Semantic descriptions are added only
after clustering by comparing assignments with independently generated
reasoning-function annotations and by inspecting representative examples.
Among 20,366 sentences with exact alignment to the semantic annotations, the
normalised mutual information between K-means assignments and semantic labels
is 0.131. We therefore use numeric cluster identities in the quantitative
analysis rather than treating the clusters as semantic categories.

\paragraph{Cluster frequencies around commitment.}
The cluster distribution changes substantially after action commitment.
$C_{11}$, $C_2$, and $C_7$ increase in frequency by 12.8, 6.3, and 5.6
percentage points, respectively, while $C_5$, $C_6$, $C_0$, and $C_1$
decrease by 9.8, 7.4, 4.4, and 2.7 percentage points. Post hoc semantic
annotations associate the increasing clusters more strongly with solving and
answering behaviour and the decreasing clusters more strongly with reading and
reconstructing the environment.

To control for ordinary progress through the reasoning trace, we compare the
cluster at each true commitment sentence with a position-matched sentence from
the same decision. The largest difference occurs for $C_{10}$, which appears
at 14.0\% of commitment sentences compared with 5.9\% of matched controls.
$C_9$ and $C_2$ show smaller excesses of 3.2 and 2.5 percentage points.

\begin{table}[htbp]
\centering
\small
\begin{tabular}{lrrr}
\toprule
Cluster & True commitment & Matched control & Excess \\
\midrule
$C_{10}$ & 14.0\% & 5.9\% & $+8.2$ pp \\
$C_9$    & 11.6\% & 8.4\% & $+3.2$ pp \\
$C_2$    & 9.3\%  & 6.8\% & $+2.5$ pp \\
$C_5$    & 15.1\% & 13.5\% & $+1.6$ pp \\
$C_7$    & 5.5\%  & 4.5\% & $+1.0$ pp \\
\bottomrule
\end{tabular}
\caption{
Frequency of selected activation clusters at true commitment sentences and
position-matched control sentences from the same decisions.
}
\label{tab:commitment-clusters-app}
\end{table}

The clusters also differ in how action confidence evolves. $C_{10}$ is
associated with the largest average increase in probability assigned to the
eventual action, 0.0245, together with an increase of 0.0400 in confidence and
0.0402 in action margin. In contrast, prefixes assigned to $C_{11}$ already
place a mean probability of 0.933 on the eventual action and show smaller
subsequent changes.

\paragraph{Clusters at action revisions.}
Approximately 4.59\% of reasoning sentences contain at least one change in
preferred action. Relative to a within-decision shuffled-event control,
$C_6$, $C_{10}$, $C_9$, and $C_5$ are enriched at these events by factors of
1.65, 1.64, 1.41, and 1.33, respectively. All four remain significant after
false-discovery-rate correction with $q=0.0024$.

The same clusters are enriched when action quality changes in either direction.
For optimal-to-suboptimal transitions, the enrichment factors for
$C_6$, $C_{10}$, $C_9$, and $C_5$ are 2.07, 1.48, 1.45, and 1.26.
For suboptimal-to-optimal transitions, they are 1.77, 1.59, 1.44, and 1.35.
This symmetry suggests that these representational states are associated with
decision revision rather than specifically with detecting or correcting an
error.

\begin{table}[htbp]
\centering
\small
\begin{tabular}{lccc}
\toprule
Cluster & Action change & Optimal to suboptimal & Suboptimal to optimal \\
\midrule
$C_6$    & $1.65\times$ & $2.07\times$ & $1.77\times$ \\
$C_{10}$ & $1.64\times$ & $1.48\times$ & $1.59\times$ \\
$C_9$    & $1.41\times$ & $1.45\times$ & $1.44\times$ \\
$C_5$    & $1.33\times$ & $1.26\times$ & $1.35\times$ \\
\bottomrule
\end{tabular}
\caption{
Enrichment of sentence-level activation clusters at changes in preferred
action and preferred-action optimality relative to within-decision
shuffled-event controls.
}
\label{tab:reasoning-events-app}
\end{table}


\section{Additional Reasoning-Text Intervention Results}
\label{app:text-interventions}

\subsection{Whole-Chain Reasoning Patching}
\label{app:whole-chain-patching}

To test whether the reasoning trajectory itself influences the current action
preference, we pair the prompt from a decision \(A\) with a reasoning prefix
from a different decision \(B\) whose emitted action differs from that of
\(A\). We evaluate prefixes taken at approximately 10\%, 50\%, and 90\% of
\(B\)'s reasoning trace. The analysis contains 396 mismatched prompt--reasoning
pairs and 396 congruent controls, each evaluated in an independent forward
pass.

Across all three cut positions, the forced-action distribution follows the
action associated with the transplanted reasoning more strongly than the action
associated with the original prompt. The probability assigned to \(a_B\) lies
between 0.598 and 0.673, whereas the probability assigned to \(a_A\) lies
between 0.129 and 0.165. At the late cut, the forced top action agrees with
the reasoning-chain action in 98.5\% of congruent pairs and 68.9\% of
mismatched pairs. The corresponding mean probabilities assigned to the
reasoning-chain action are 0.971 and 0.673.

\begin{figure}[t]
\centering
\includegraphics[width=0.62\textwidth]{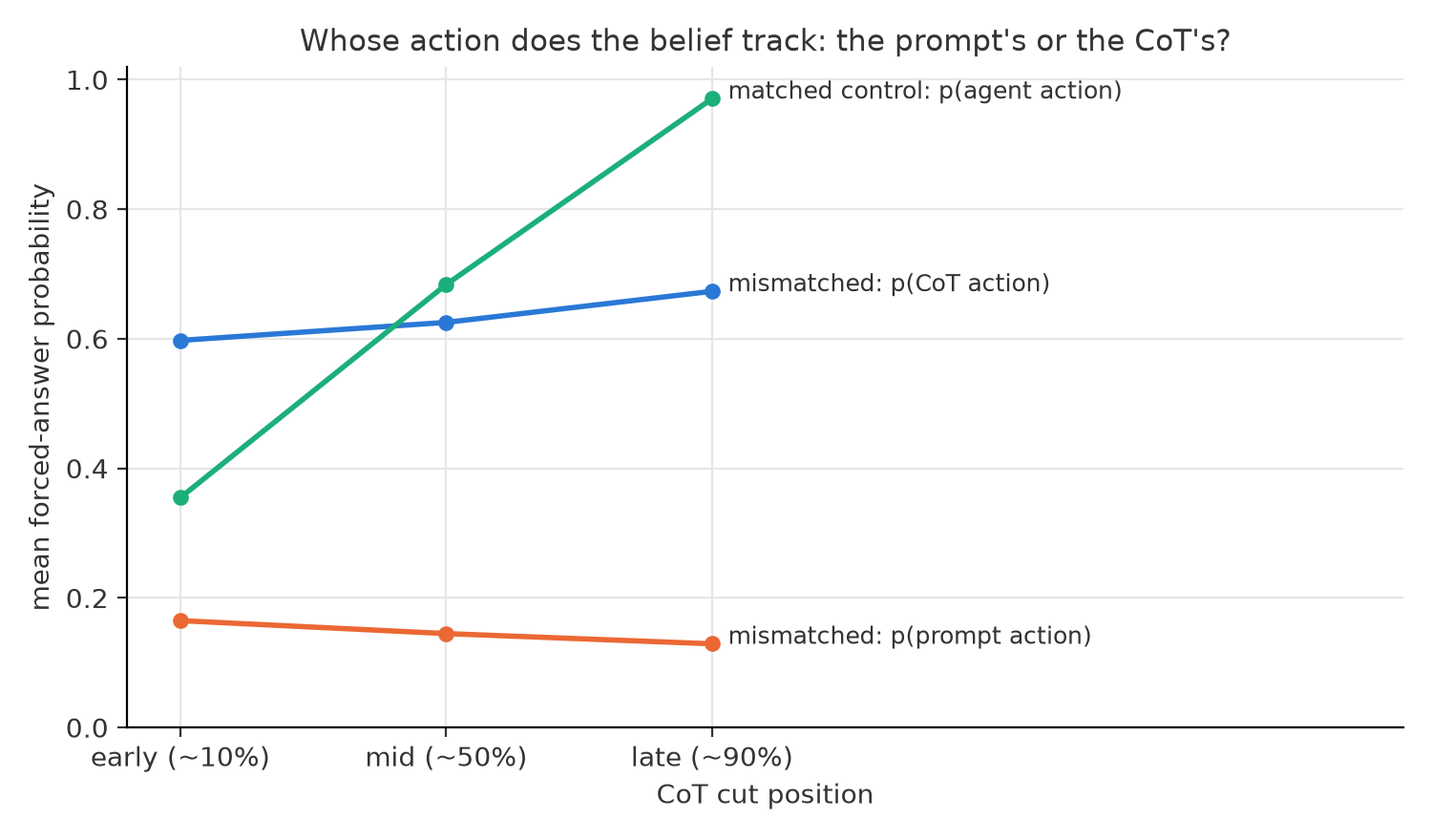}
\caption{
Whole-chain reasoning-text patching. The prompt from decision \(A\) is paired
with a reasoning prefix from decision \(B\), whose emitted action differs from
that of \(A\). Across early, middle, and late cuts, the forced-action
distribution preferentially follows the transplanted reasoning rather than the
original prompt. The experiment uses 396 mismatched pairs and 396 congruent
controls.
}
\label{fig:cot-swap-app}
\end{figure}

\subsection{Single-Sentence Interventions}
\label{app:sentence-patching}

We next test whether a local textual intervention is sufficient to redirect the
current action preference. We sample 240 reasoning prefixes across six bins of
the current eventual-action probability
\(c_{t,k}=\pi_{t,k}(a_t)\), with 40 prefixes per bin and at most three prefixes
from any one decision. At each prefix, we append one of three sentences:

\begin{enumerate}
    \item a counter-evidence sentence asserting that the current runner-up
    action is correct;
    \item a neutral sentence of similar form that names no action;
    \item a supportive sentence asserting that the current top action is
    correct.
\end{enumerate}

This yields 720 independent forced-action evaluations. The counter-evidence
template is
``Wait, I should double-check this. Looking at the grid again, the correct move
here is \textsc{action}.''

For the main analysis, we restrict to the 201 cases in which the injected
runner-up action differs from the action eventually emitted by the model.
The forced top action switches to the injected action in 100\% of cuts with
\(c_{t,k}<0.7\), 92.5\% for
\(0.7\leq c_{t,k}<0.85\), 90.0\% for
\(0.85\leq c_{t,k}<0.95\), and 62.5\% for
\(c_{t,k}\geq0.95\). Susceptibility decreases with current action confidence
(Spearman \(\rho=-0.421\), \(p=5.0\times10^{-10}\)).

Action confidence is correlated with reasoning position
(\(\rho=0.538\)). The intervention effect also declines with progress through
the trace (\(\rho=-0.436\)), but action confidence retains a smaller
independent association after controlling for position
(partial \(\rho=-0.175\), \(p=0.013\)).

The neutral sentence changes the preferred action in 9.6\% of cuts overall,
while the supportive sentence increases the mean probability assigned to the
current top action from 0.757 to 0.989. These controls indicate that the
counter-evidence effect is primarily driven by the semantic content of the
injected sentence rather than by appending additional text.

\begin{figure}[t]
\centering
\includegraphics[width=\textwidth]{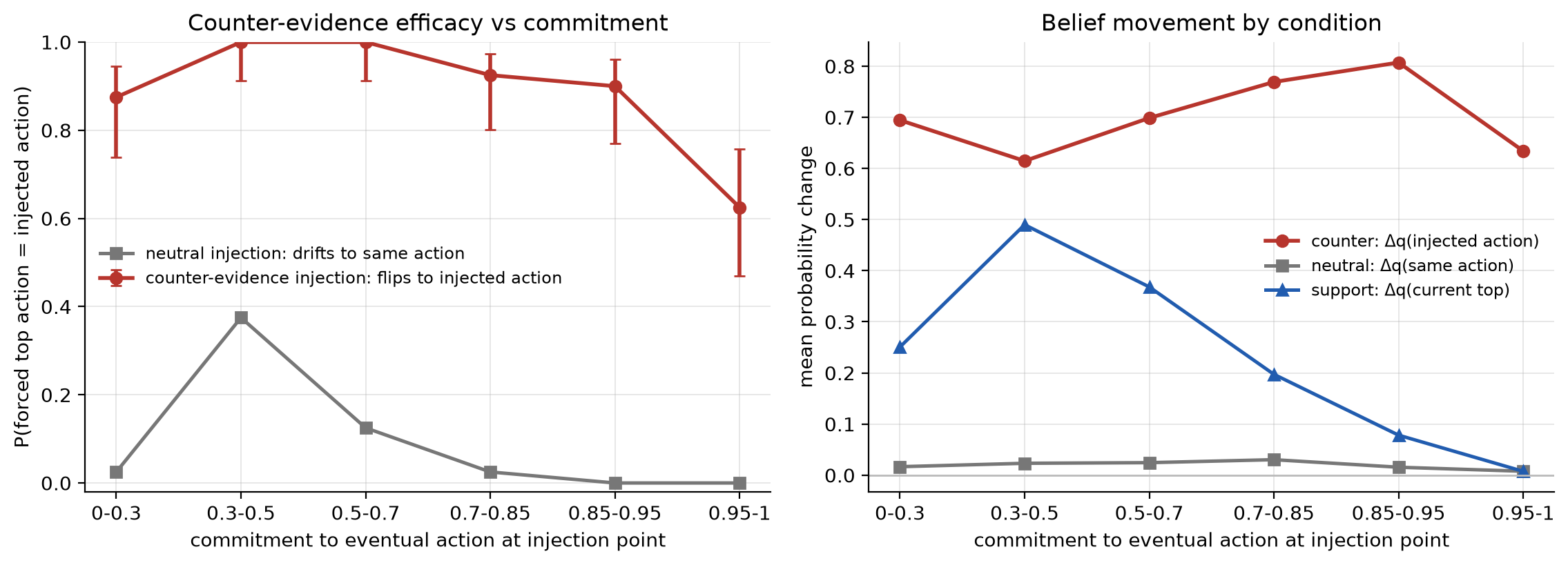}
\caption{
Single-sentence intervention results across 240 reasoning prefixes and 720
independent forced-action evaluations.
\textbf{Left:} probability that the forced top action becomes the injected
runner-up action as a function of current action confidence.
\textbf{Right:} mean probability movement of the injected action under
counter-evidence, neutral, and supportive interventions.
}
\label{fig:injection-app}
\end{figure}

\section{Additional Activation Intervention Results}
\label{app:activation-interventions}

\subsection{Commitment Readout and Intervention Setup}
\label{app:activation-setup}

We target the residual activations over the final up-to-12 reasoning tokens
before the answer-forcing suffix. A nonlinear readout trained on this
reasoning-window representation predicts inverse normalised action entropy with
held-out Spearman \(\rho=0.771\pm0.009\). For comparison, a length-oracle
baseline reaches 0.644 and a positional baseline 0.371. Evaluation uses
trajectory-grouped splits.

The interventions test whether the same representation can be manipulated to
control the current action. We distinguish between any change in the forced top
action and a change specifically to the intended target action. The main paper
summarises these results in \Cref{fig:readwrite}.

\subsection{Steering, Transplantation, and Ablation Controls}
\label{app:activation-controls}

We evaluate several intervention families at the reasoning-window site.

\paragraph{Steering.}
We add an action-associated rank-1 direction to the reasoning-window residual
state and compare it with a norm-matched random direction. At the representative
magnitude shown in \Cref{fig:readwrite}, both targeted and random steering
change the action to the intended target in 3.6\% of cases. The targeted
direction therefore does not outperform the matched random control in this
setting.

\paragraph{Transplantation.}
We replace components of the recipient reasoning-window representation with
those from a donor prefix associated with a different preferred action.
Low-rank transplantation does not produce intended flips in the tested
settings. Replacing the entire 12-token reasoning-window state is more
effective, changing the forced action in 25.0\% of donor--recipient pairs and
redirecting it to the intended donor-associated action in 20.8\%.

\paragraph{Sub-block deletion.}
Skipping an attention or MLP sub-block at one layer changes the forced action
in up to 29.2\% of cases. Because this intervention has no intended target, it
is best interpreted as a disruption control rather than directed steering.

\paragraph{Instruction control.}
Appending an explicit instruction to disregard the mismatched reasoning changes
the forced action in 5.8\% of cases and moves it to the intended target in
1.0\%.

These negative results should not be interpreted as showing that the
reasoning-window representation is causally irrelevant. The tested
interventions modify only particular layers, directions, and subspaces, and no
known positive-control steering direction is available for this model. They do
show, however, that the highly decodable representation at this site is not
reliably writable under the interventions tested.



\clearpage
  \section{Illustrative Belief--Action Decomposition}
  \label{app:belief-action-overview}

  \begin{figure}[htbp]
  \centering
  \includegraphics[width=\textwidth]{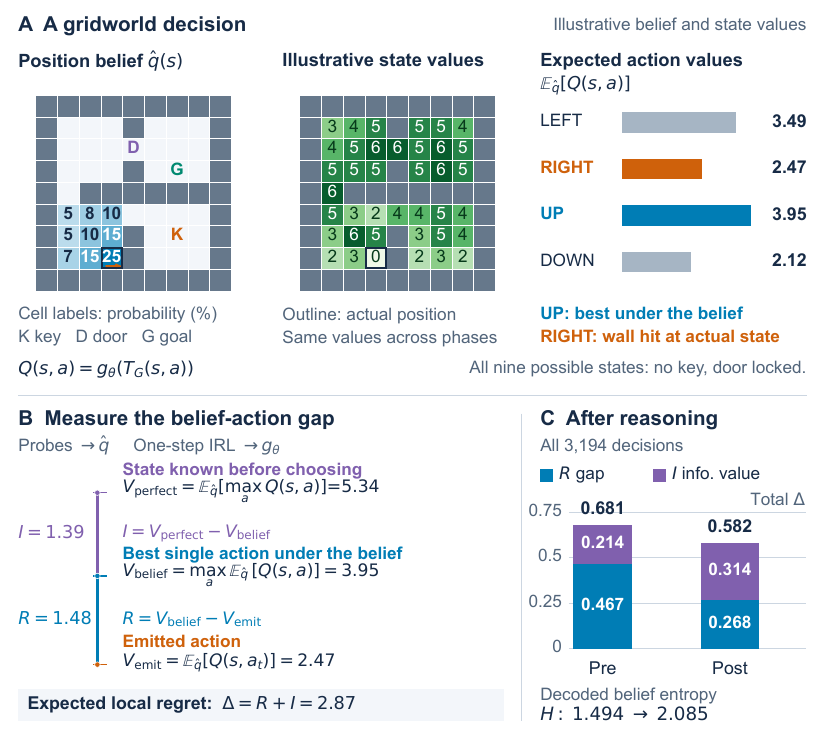}
  \caption{\textbf{Belief--action decomposition: worked example and empirical summary.}
  \textbf{(A--B)} A constructed DoorKey example assigns belief probability to nine
  positions, all with no key and a locked door; blank belief cells have zero
  probability. The green map assigns values to all 39 non-wall positions.
  For this illustration only, $g(x,y,h,d)=v(x,y)$ across valid states; the
  door-cell value applies when the door is open. With $Q(s,a)=g(T_G(s,a))$,
  the emitted RIGHT action has expected value $2.47$, the best single action
  UP has value $3.95$, and choosing after learning the state yields $5.34$.
  Thus the belief--action gap is $R=1.48$, the value of perfect state
  information is $I=1.39$, and expected local regret is $\Delta=R+I=2.87$.
  RIGHT is blocked at the outlined actual position. These beliefs, values,
  and action are illustrative, not fitted observations or task rewards.
  \textbf{(C)} Empirical endpoint means across all 3,194 decisions
  (\Cref{tab:gap-pre-post}), using representation-based beliefs and
  agent-supervised value models fitted separately before and after reasoning.
  $R$ decreases, while $I$ and decoded-belief entropy $H$ increase.}
  \label{fig:belief-action-overview}
  \end{figure}

\end{document}

%% file: math_commands.tex
\usepackage{amsmath,amsfonts,bm}

\def\eqref#1{equation~\ref{#1}}

\def\1{\bm{1}}

\DeclareMathAlphabet{\mathsfit}{\encodingdefault}{\sfdefault}{m}{sl}
\SetMathAlphabet{\mathsfit}{bold}{\encodingdefault}{\sfdefault}{bx}{n}



%% file: macros.tex
\usepackage[capitalize,noabbrev]{cleveref}
\crefname{section}{\S}{\S\S}
\crefname{table}{Tab.}{Tab.}
\crefname{figure}{Fig.}{Figs.}
\crefname{algorithm}{Alg.}{}
\crefname{equation}{Eq.}{Eq.}
\crefname{appendix}{App.}{App.}
\crefname{theorem}{Theorem}{}
\crefname{restatableTheorem}{Theorem}{}
\crefname{prop}{Proposition}{}
\crefname{definition}{Def.}{}
\crefname{cor}{Corollary}{}
\crefname{observation}{Observation}{}
\crefname{assumption}{Assumption}{}
\crefname{hyp}{Hyp.}{Hypotheses}
\crefformat{section}{\S#2#1#3}
\crefname{namedtheorem}{Hyp.}{Hypotheses}

\usepackage[textsize=tiny]{todonotes}

\definecolor{Blue}{RGB}{33,92,175}   
\definecolor{Green}{RGB}{98,115,19}      
\definecolor{PurpleDark}{RGB}{140,10,89} 
\definecolor{Purple}{RGB}{163,7,116} 
\definecolor{Gray}{RGB}{111,111,111} 
\definecolor{Red}{RGB}{183,53,45}   
\definecolor{Petrol}{RGB}{0,120,148}
\definecolor{Bronze}{RGB}{142,103,19} 
\definecolor{Orange}{RGB}{230, 100, 50}